\documentclass[conference]{IEEEtran}
\IEEEoverridecommandlockouts
\usepackage{cite}
\usepackage{amsmath,amssymb,amsfonts}
\usepackage{algorithmic}
\usepackage{graphicx}
\usepackage{tikz}
\usetikzlibrary{arrows.meta,calc}
\usepackage{stfloats}
\usepackage{textcomp}
\usepackage{xcolor}
\usepackage{booktabs}
\usepackage{array}
\usepackage[framemethod=TikZ]{mdframed}
\usepackage{needspace}
\usepackage{url}
\usepackage[hidelinks]{hyperref}

\def\BibTeX{{\rm B\kern-.05em{\sc i\kern-.025em b}\kern-.08em
    T\kern-.1667em\lower.7ex\hbox{E}\kern-.125emX}}
\definecolor{missingred}{RGB}{176,0,32}
\definecolor{missingbg}{RGB}{255,244,246}
\definecolor{flagblue}{RGB}{0,72,140}
\definecolor{flagbg}{RGB}{240,246,255}

\begin{document}
\title{Model-Based Trailer-Sway\\
Suppression for Autonomous Passenger Vehicles}
\author{\IEEEauthorblockN{Alec McFadyen\textsuperscript{*}, Pablo Covarrubias De La Torre\textsuperscript{*}, Connor McGovern\textsuperscript{*}}
\IEEEauthorblockA{Boston, Massachusetts\\
alecmcfadyen9151@gmail.com, pcorrtorr@gmail.com, connorm7788@icloud.com}
}
\maketitle
\begingroup
\renewcommand\thefootnote{*}
\footnotetext{All authors contributed equally to this work.}
\endgroup
%=====================================================================
\begin{abstract}
Trailer sway is a persistent contributor to severe crashes among light
passenger vehicles towing trailers, yet it remains largely unaddressed by
the autonomous vehicle systems now entering this new era of automation.
National fatal crash data record 146 fatal crashes involving jackknifing (when a trailer oscillates and swings around the tow vehicle, forming an acute angle with it)
among light passenger vehicles towing trailers between 2020 and 2024\cite{farsjackknifing},
excluding large trucks; because jackknifing is coded as a discrete
terminal event, this figure represents a lower bound on fatalities
attributable to trailer instability. Improper cargo placement rearward of
the trailer axle reduces, and can invert, the tongue load transmitted
through the hitch. The resulting loss of vertical load at the tow
vehicle's rear axle diminishes the lateral tire force available there and
degrades yaw stability of the coupled system, admitting a divergent
lateral oscillation of the trailer known as trailer sway. Under adverse
combinations of speed and loading, this oscillation can progress to
jackknifing, in which the articulation angle diverges rapidly and
directional control is lost. This work develops a linear state-space
model of the yaw-plane dynamics of a light vehicle--trailer combination,
parameterized to expose tongue load as an explicit stability-determining
quantity. From this model a linear quadratic regulator is synthesized to suppress articulation-angle divergence. Because consumer trailers carry no sensors, the two trailer states the controller needs, trailer yaw rate and articulation angle, are estimated by a Kalman filter from
measurements taken only on the tow vehicle. The controller is then
implemented and evaluated in a MuJoCo \cite{mujoco} multibody simulation across a range of longitudinal speeds and tongue-load conditions, including the
negative-tongue-load regime. Results show a significant decrease in
trailer oscillation when controlling a given vehicle's steering with an LQR using
tow-vehicle IMU and LiDAR data. These findings indicate that LQR controlled
anti-sway systems are a viable control method for future autonomous
vehicles.
\end{abstract}
%=====================================================================
\section{Introduction}
Over the past several years, autonomous vehicle prevalence and their use
cases have continued to expand. However, one overlooked area has been potential issues that arise during self-driving truck operation. One example
is moving items by towing a trailer, due to how the trailer
changes the vehicle's inertia and response to simple steering movements,
leaving autonomous vehicles especially prone to lateral instability and a
phenomenon known as trailer sway or ``fishtailing.'' In 2024, approximately 361,128 vehicles towing trailers in the US
got into accidents\cite{traileraccidents}, many of which are caused by fishtailing.
Within the autonomous vehicle industry, an increasing focus is placed on
the trucking industry and large tractor-semitrailer AVs. Companies like
Aurora Innovation and Kodiak Robotics have begun production and
distribution of trucks capable of SAE Level 4 autonomy. Although these larger semi-trucks are excellent for shipping goods, they
depend on extensive sensor suites. The Aurora Driver combines
high-resolution long-range cameras, imaging radar, and Aurora's own
frequency-modulated continuous-wave (FMCW) FirstLight lidar
\cite{aurora10k}, while Kodiak's fifth-generation truck carries 18
sensors, including four lidars, four radars, and ten cameras, mounted in
pods that replace the side mirrors \cite{kodiakgen5}. This hardware, along
with the narrow use cases these trucks are built for, puts it out of reach
for everyday consumers using trailers.

Furthermore, through talking with Gustavo Perez, a representative working at International Trucking (Gustavo Perez, personal communication, July 2026), we learned that even major trucking companies still have difficulties accounting for sudden incidents of instability and trailer sway caused by outside forces such as wind or a bumpy road\cite{disturbancesources}.

However, sensors are not the only difficulty with modern-day control
algorithms used by semi-trucks to prevent trailer sway. Most rely on
electronically controlled braking systems (ECBS) for stability and sway
control, alongside expensive suspension systems, and often require
specific packaging scenarios and weight distributions. Consumer trailers
do not have these suspension and braking systems; additionally, many
users are not aware of how to pack their items and distribute weight
properly. We address this problem by developing a novel sway countering controller, which we extensively test in a realistic trailer simulation, and which requires no specialized hardware or knowledge from the user. The trailer itself is left uninstrumented, so the controller infers the trailer's yaw rate and its angle relative to the tow vehicle from sensors the tow vehicle already carries.
%=====================================================================
\section{Problem Definition}
\subsection{Trailer Sway as a Dynamic Instability}
When an autonomous vehicle pulls a trailer, wind, bumps,
or quick turns can cause the trailer to swing in a dangerous
oscillation known as trailer sway or fishtailing. If the vehicle
cannot automatically stabilize this movement and prevent it
under real-world road conditions, it risks losing control, caus-
ing an accident, or forcing the system to shut down. Solving
this is critical because trailer sway naturally gets worse over
time and can quickly spiral out of control and cause fatal
accidents. Since an autonomous system lacks human “feel,” it
must be able to recognize and stop these dangerous oscillations
on its own to keep the vehicle safely on the road. The scope of this work is deliberately narrow where a successful system must protect the vehicle from crashing while ensuring
the ride stays within the confines of the lane as efficiently as
possible for multiple speeds and trailer distributions while only using sensors mounted on the towing vehicle and only using steering as the method of control.

Solving this is critical because trailer sway naturally gets worse over
time and can quickly spiral out of control and cause fatal accidents.
Since an autonomous system lacks human ``feel,'' it must be able to
recognize and stop these dangerous oscillations on its own to keep the
vehicle safely on the road.
A successful system must protect the vehicle from crashing while ensuring
the ride stays within the confines of the lane as efficiently as possible
for multiple speeds and trailer distributions.

\subsection{Related Work}
In our literature review, we found that most existing methods for
countering trailer sway were designed for human-driven vehicles and rely
on braking. Some brake only the trailer, slowing it so that it is pulled
back in line, while others brake individual wheels on the car and trailer
to create a yaw moment that opposes the sway
\cite{ieee11397333,plos2022,wu2017,sciencedirect2019}. Commercial versions
of these systems run on the tow vehicle's electronic stability control and
read its wheel speed, yaw rate, lateral acceleration, steering angle, and
brake pressure sensors \cite{boschtsm}. Because they need hardware that
can brake individual wheels, they cannot be used on vehicles or trailers
without it. Steering-based sway mitigation has also been proposed
\cite{boschsteer}, but nearly all prior work assumes a human driver who
remains responsible for keeping the vehicle in its lane. An autonomous
vehicle must handle both tasks at once, damping sway while staying in the
lane, and this combined problem is the one we address.
%=====================================================================
\section{Mathematical Model}
\subsection{Modeling Assumptions}
To create the state space model we made several assumptions:
\begin{itemize}
  \item Tire behavior is linear, so tire force is proportional to slip
        angle, $F_y = -C\alpha$. Standard for bicycle-model analysis
        \cite{openvd} and accurate at the small slip angles of
        near-straight travel.
  \item Steer angle, sideslip, and articulation angle are small enough
        to use $\cos x \approx 1$ and $\sin x \approx x$. Stays within
        1.5\% and 0.5\% of the true values below 10 degrees, and the
        controlled articulation angle stays under about 5 degrees.
  \item Forward speed is constant, so $V$ is a parameter rather than a
        state. Fits a steering-only controller with no braking or
        throttle, where speed changes slowly compared with the
        sub-second sway period.
  \item Motion is planar, so pitch, roll, heave, and suspension are
        dropped. Keeps the model in the yaw plane, where sway happens,
        and leaves load-transfer effects to the MuJoCo simulation.
  \item The hitch is a frictionless barrel hinge with one degree of
        freedom. Captures the only hitch rotation that matters for yaw,
        and ignoring friction makes the model slightly pessimistic.
  \item The road is flat and straight with a consistent surface. Allows
        for keeping the straight-line scope and simplifying the problem.
\end{itemize}
These simplifications are what make the stability question solvable in
closed form rather than only numerically, and that is the reason they
matter here. We later revisit several of the more restrictive ones using
a nonlinear MuJoCo simulation, which lets us check the linear predictions
against a model that does not require small angles or planar motion.
\subsection{Vehicle Parameters}
The results within this section of the paper utilize the measurements
that we approximated for our scaled Neoracer test platform, listed in
Table~\ref{tab:params}. We ran every case with two different trailer
configurations, with only a difference in distance to trailer CG from
hitch, $a_2$. In the baseline configuration the trailer CG sits
$0.067$~m ahead of the trailer axle, and in the destabilized
configuration it sits essentially on top of the axle, $0.0015$~m ahead of
it, which drives the static tongue load to almost zero.
\begin{table}[t]
\caption{Vehicle parameters used for the scaled Neoracer platform, as
coded in \texttt{swaykalmaanboundedgraph.jl}. Entries marked
``est.''\ are estimates rather than measurements; entries marked
``exact'' were computed from the CAD geometry.}
\label{tab:params}
\centering
\scriptsize
\setlength{\tabcolsep}{3pt}
\begin{tabular}{@{}llll@{}}
\toprule
Symbol & Quantity & Value & Source \\
\midrule
$m_1$ & tractor (car) mass            & $5.5$ kg              & measured \\
$I_1$ & tractor yaw inertia           & $0.053$ kg\,m$^2$     & est. \\
$a$   & tractor CG $\to$ front axle   & $0.144$ m             & wheelbase/2 \\
$b$   & tractor CG $\to$ rear axle    & $0.144$ m             & wheelbase/2 \\
$h$   & rear axle $\to$ hitch         & $0.042$ m             & CAD \\
$C_f$ & front cornering stiffness     & $300$ N/rad           & est. \\
$C_r$ & rear cornering stiffness      & $300$ N/rad           & est. \\
$m_2$ & trailer mass (frame + wheels + cargo) & $4.4$ kg      & exact \\
$L_2$ & hitch $\to$ trailer axle      & $0.2475$ m            & CAD (fixed) \\
$C_2$ & trailer cornering stiffness   & $100$ N/rad           & est. \\
\midrule
\multicolumn{4}{@{}l}{\textit{Baseline (forward CG) configuration}} \\
$I_2$ & trailer yaw inertia           & $0.0230$ kg\,m$^2$    & exact \\
$a_2$ & hitch $\to$ trailer CG        & $0.181$ m             & exact \\
\midrule
\multicolumn{4}{@{}l}{\textit{Destabilized (rearward CG) configuration}} \\
$I_2$ & trailer yaw inertia           & $0.02107$ kg\,m$^2$   & exact \\
$a_2$ & hitch $\to$ trailer CG        & $0.27866$ m           & exact \\
\bottomrule
\end{tabular}
\end{table}
The trailer mass, hitch distance, and weight distribution were not
measured directly, but were instead extrapolated from the Neoracer's own
parameters, scaled by the average mass ratio between production electric
vehicles and the utility trailers they tow. Taking a Rivian R1 at
6{,}850~lb against an average utility trailer at 1{,}500~lb unloaded and
3{,}000~lb loaded gives mass ratios of $4.57$ and $2.28$ respectively;
applied to the Neoracer's $5.5$~lb this yields a scaled trailer of
$1.20$~lb unloaded and $2.41$~lb loaded.

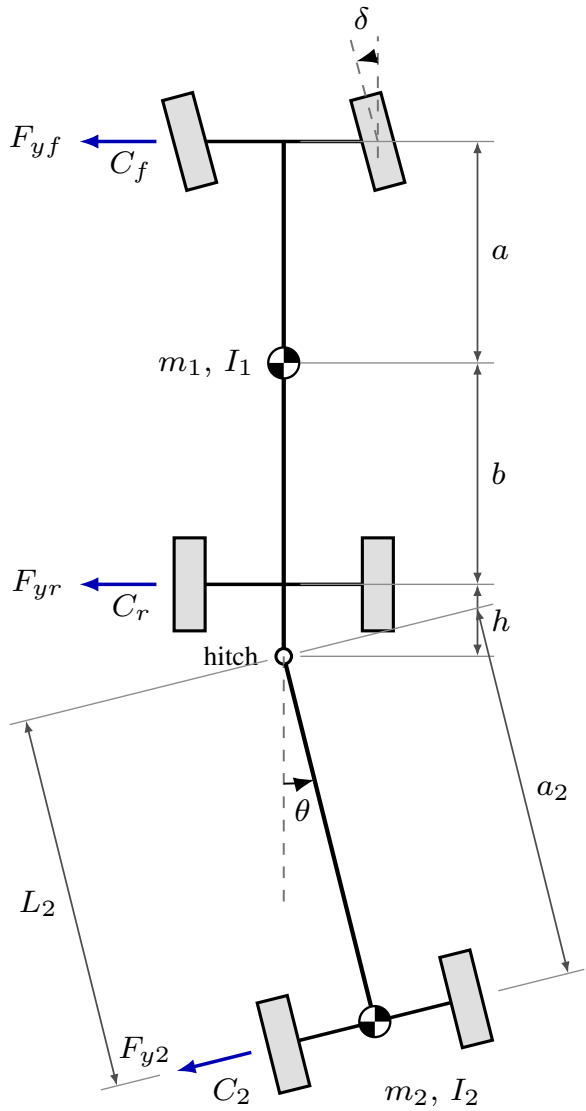
\begin{figure}[!t]
\centering
\resizebox{0.9\columnwidth}{!}{%
\begin{tikzpicture}[
    >=Latex,
    every node/.style={font=\footnotesize},
    spine/.style={draw=black,line width=1.1pt},
    axleline/.style={draw=black,line width=0.9pt},
    wheel/.style={draw=black,fill=gray!25,line width=0.8pt},
    dim/.style={draw=black!70,line width=0.45pt,{Latex[length=3pt]}-{Latex[length=3pt]}},
    ext/.style={draw=black!45,line width=0.35pt},
    ref/.style={draw=black!55,line width=0.5pt,dashed},
    force/.style={draw=blue!70!black,line width=0.9pt,-{Latex[length=5pt]}},
    cg/.pic={
      \fill[white] (0,0) circle (0.14);
      \fill[black] (0,0) -- (0.14,0) arc (0:90:0.14) -- cycle;
      \fill[black] (0,0) -- (-0.14,0) arc (180:270:0.14) -- cycle;
      \draw[line width=0.6pt] (0,0) circle (0.14);
    },
]
% ---------- geometry (scaled drawing units) ----------
\def\tw{0.85}   % half track width, same for every axle
\def\ww{0.14}   % half wheel width
\def\wl{0.42}   % half wheel length
\def\dlt{15}    % drawn steer angle
\def\tht{14}    % drawn articulation angle

\coordinate (Faxle) at (0,4.0);
\coordinate (CG)    at (0,2.0);
\coordinate (Raxle) at (0,0);
\coordinate (Hitch) at (0,-0.65);

% ================= TRACTOR =================
\draw[spine] (Faxle) -- (Hitch);

% front axle, wheels steered by delta about their own centres
\draw[axleline] ($(Faxle)+(-\tw,0)$) -- ($(Faxle)+(\tw,0)$);
\foreach \s in {-1,1}{
  \begin{scope}[shift={($(Faxle)+(\s*\tw,0)$)},rotate=\dlt]
    \draw[wheel] (-\ww,-\wl) rectangle (\ww,\wl);
  \end{scope}
}
% steer-angle reference and arc on right front wheel
\coordinate (RF) at ($(Faxle)+(\tw,0)$);
\draw[ref] ($(RF)+(0,-0.15)$) -- ($(RF)+(0,0.95)$);
\draw[ref] (RF) -- ($(RF)+({90+\dlt}:0.95)$);
\draw[->,line width=0.5pt] ($(RF)+(0,0.75)$) arc (90:{90+\dlt}:0.75);
\node at ($(RF)+({90+\dlt/2}:1.12)$) {$\delta$};

% rear axle
\draw[axleline] ($(Raxle)+(-\tw,0)$) -- ($(Raxle)+(\tw,0)$);
\foreach \s in {-1,1}
  \draw[wheel] ($(Raxle)+(\s*\tw-\ww,-\wl)$) rectangle ($(Raxle)+(\s*\tw+\ww,\wl)$);

% tyre forces (lateral, at left wheels)
\draw[force] ($(Faxle)+(-\tw-0.3,0)$) -- ($(Faxle)+(-\tw-1.0,0)$)
  node[left] {$F_{yf}$};
\draw[force] ($(Raxle)+(-\tw-0.3,0)$) -- ($(Raxle)+(-\tw-1.0,0)$)
  node[left] {$F_{yr}$};
\node[below left,inner sep=1pt] at ($(Faxle)+(-\tw-0.3,-0.05)$) {$C_f$};
\node[below left,inner sep=1pt] at ($(Raxle)+(-\tw-0.3,-0.05)$) {$C_r$};

% tractor CG
\pic at (CG) {cg};
\node[left=4pt] at (CG) {$m_1,\, I_1$};

% hitch
\filldraw[fill=white,line width=0.8pt] (Hitch) circle (0.07);
\node[left=3pt,font=\scriptsize] at (Hitch) {hitch};

% dimensions a, b, h (right side)
\def\dx{1.75}
\foreach \p in {Faxle,CG,Raxle,Hitch}
  \draw[ext] ($(\p)+(0.15,0)$) -- ($(\p)+(\dx+0.15,0)$);
\draw[dim] ($(Faxle)+(\dx,0)$) -- node[right] {$a$} ($(CG)+(\dx,0)$);
\draw[dim] ($(CG)+(\dx,0)$)    -- node[right] {$b$} ($(Raxle)+(\dx,0)$);
\draw[dim] ($(Raxle)+(\dx,0)$) -- node[right] {$h$} ($(Hitch)+(\dx,0)$);

% ================= TRAILER (rotated by theta) =================
\draw[ref] (Hitch) -- ($(Hitch)+(0,-2.3)$);
\begin{scope}[shift={(Hitch)},rotate=\tht]
  \coordinate (Taxle) at (0,-3.4);
  \coordinate (TCG)   at (0,-3.4);
  \draw[spine] (0,-0.07) -- (Taxle);
  \draw[axleline] ($(Taxle)+(-\tw,0)$) -- ($(Taxle)+(\tw,0)$);
  \foreach \s in {-1,1}
    \draw[wheel] ($(Taxle)+(\s*\tw-\ww,-\wl)$) rectangle ($(Taxle)+(\s*\tw+\ww,\wl)$);
  \pic at (TCG) {cg};

  \draw[force] ($(Taxle)+(-\tw-0.3,0)$) -- ($(Taxle)+(-\tw-1.0,0)$)
    coordinate (Fy2end);
  \coordinate (C2lab) at ($(Taxle)+(-\tw-0.55,-0.3)$);

  % L2 on left side, a2 on right side
  \draw[ext] (-0.15,0) -- (-2.55,0);
  \draw[ext] ($(Taxle)+(-\tw-1.15,0)$) -- ($(Taxle)+(-2.55,0)$);
  \draw[dim] (-2.4,0) -- (-2.4,-3.4) coordinate[midway] (L2lab);
  \draw[ext] (0.15,0) -- (1.95,0);
  \draw[ext] ($(TCG)+(\tw+0.3,0)$) -- ($(TCG)+(1.95,0)$);
  \draw[dim] (1.8,0) -- (1.8,-3.4) coordinate[midway] (a2lab);
  \coordinate (m2lab) at ($(TCG)+(0.35,-0.75)$);
\end{scope}
\node[above left,inner sep=1pt] at (Fy2end) {$F_{y2}$};
\node at (C2lab) {$C_2$};
\node[left] at (L2lab) {$L_2$};
\node[right] at (a2lab) {$a_2$};
\node at (m2lab) {$m_2,\, I_2$};

% articulation angle
\draw[->,line width=0.5pt] ($(Hitch)+(0,-1.15)$) arc (270:{270+\tht}:1.15);
\node at ($(Hitch)+({270+\tht/2}:1.4)$) {$\theta$};

\end{tikzpicture}
}
\caption{Schematic of the coupled tractor--trailer bicycle model in the
destabilized (rearward-CG) loading configuration, showing the front, rear
and trailer tire forces $F_{yf}$, $F_{yr}$ and $F_{y2}$, the steer angle
$\delta$, the articulation angle $\theta$, and the segment lengths $a$,
$b$, $h$, $L_2$ and $a_2$. Numerical values for every labelled quantity
are given in Table~\ref{tab:params}.}
\label{fig:schematic}
\end{figure}

Fig.~\ref{fig:schematic} shows the corresponding tractor--trailer
schematic, defining the sign conventions for the steer angle $\delta$
and articulation angle $\theta$ and for the front, rear, and trailer
tire forces $F_{yf}$, $F_{yr}$, and $F_{y2}$ used throughout the
derivation below.
\subsection{Derivation of the Four-State Model}
To build the four-state model, we treated the tractor and the trailer as
two separate rigid bodies moving in the plane, connected to each other by
a constraint force acting through the hitch. Applying Newton's second law
in the lateral direction, and about the vertical axis of each body, gives
four equations of motion. Rather than solving for the hitch force
directly and eliminating it afterward, we used the hitch constraint to
write the trailer's motion in terms of the tractor's own states, which
lets us collapse both bodies into a single mass matrix.
Using $q = [v_1,\; r_1,\; r_2]^\mathsf{T}$ for the three dynamic states,
this reduces to
\begin{equation}
M \dot{q} = T F + G x
\label{eq:mqdot}
\end{equation}
where $F = [F_{yf},\; F_{yr},\; F_{y2}]^\mathsf{T}$ collects the three
tire forces (front, rear, and trailer), $M$ is the generalized mass
matrix, $T$ resolves those tire forces into a net lateral force and two
yaw moments, and $G$ collects the centripetal terms that show up because
we are writing the dynamics in the tractor's rotating body frame rather
than a fixed inertial one.
\begin{equation}
M =
\begin{bmatrix}
m_1 + m_2      & -m_2 (b+h)           & -m_2 a_2 \\[2pt]
-m_2 (b+h)     & I_1 + m_2 (b+h)^2    & m_2 a_2 (b+h) \\[2pt]
-m_2 a_2       & m_2 a_2 (b+h)        & I_2 + m_2 a_2^2
\end{bmatrix}
\label{eq:M}
\end{equation}
\begin{equation}
T =
\begin{bmatrix}
1 & 1  & 1 \\[2pt]
a & -b & -(b+h) \\[2pt]
0 & 0  & -L_2
\end{bmatrix}
\end{equation}
\begin{equation}
G =
\begin{bmatrix}
0 & -(m_1+m_2)V & 0 & 0 \\[2pt]
0 & m_2 (b+h) V & 0 & 0 \\[2pt]
0 & m_2 a_2 V   & 0 & 0
\end{bmatrix}
\label{eq:TG}
\end{equation}
Every off-diagonal term in $M$ is proportional to the trailer mass $m_2$.
That is really just a mathematical way of saying that a heavier trailer,
or one whose center of gravity sits further from the hitch, couples more
tightly into the tractor's own motion, which matters later when we look
at how sensitive the system is to loading.
The tire forces themselves come from the Linear cornering stiffness model we 
introduced earlier, $F_{y,i} = -C_i \alpha_i$, with the slip angle at each axle
following from the small-angle kinematics we assumed. Writing this out in
terms of the states and the steering input gives
\begin{equation}
F = C_t x + C_\delta \delta
\label{eq:tireforce}
\end{equation}
\begin{figure*}[!t]
\normalsize
\begin{equation}
C_t =
\begin{bmatrix}
-\dfrac{C_f}{V} & -\dfrac{C_f a}{V}    & 0                    & 0 \\[8pt]
-\dfrac{C_r}{V} & \dfrac{C_r b}{V}     & 0                    & 0 \\[8pt]
-\dfrac{C_2}{V} & \dfrac{C_2 (b+h)}{V} & \dfrac{C_2 L_2}{V}   & -C_2
\end{bmatrix}
\qquad\qquad
C_\delta =
\begin{bmatrix} C_f \\[6pt] 0 \\[6pt] 0 \end{bmatrix}
\label{eq:CtCd}
\end{equation}
\hrulefill
\vspace*{4pt}
\end{figure*}
The last row of $C_t$ is the one that actually produces sway. The
trailer's slip angle depends directly on the articulation angle $\theta$,
and on both yaw rates through the vehicle geometry, so a trailer that is
already rotated relative to the tractor generates a force that feeds back
into its own rotation. Whether that feedback works to correct the trailer
or to make things worse is exactly the question the eigenvalues answer in
the next section.
The fourth state, articulation angle, is not a dynamic state in the same
sense as the other three. It is defined kinematically as the difference
between the tractor's heading and the trailer's heading, so
\begin{equation}
\dot{\theta} = r_1 - r_2
\label{eq:thetadot}
\end{equation}
Adding this equation to the three we got from inverting the mass matrix
gives us the full four-state system
\begin{equation}
x = [v_1,\; r_1,\; r_2,\; \theta]^\mathsf{T},
\qquad
\dot{x} = A x + B \delta
\label{eq:fourstate}
\end{equation}
with
\begin{equation}
A =
\begin{bmatrix}
M^{-1}\!\left(T C_t + G\right) \\[4pt]
\begin{matrix} 0 & 1 & -1 & 0 \end{matrix}
\end{bmatrix},
\qquad
B =
\begin{bmatrix}
M^{-1} T C_\delta \\[4pt] 0
\end{bmatrix}
\label{eq:AB4}
\end{equation}
where the top three rows are the dynamics recovered by inverting the mass
matrix and the last row is the articulation kinematics of
(\ref{eq:thetadot}). Numerical values of $A$ and $B$ at the design speed
are given in Appendix~\ref{app:matrices}.
These four states are the minimum we need to describe the problem. Two of
them describe how the tractor itself is moving, lateral velocity and yaw
rate. One describes how the trailer is rotating, trailer yaw rate. And
one describes the geometry that couples the two together, articulation
angle. Only the two tractor states can be measured on the tow vehicle; trailer
yaw rate and articulation angle cannot, and Section~\ref{sec:estimation} describes how we estimate them. Because forward speed is held constant, lateral velocity and
sideslip carry the same information, related by $v_1 = V\beta$ under the
small-angle assumption. We use $v_1$ as the state in the model itself,
but we will report eigenvector participation in terms of $\beta$ later
on, since a sideslip angle is easier to compare against the other three
states in magnitude.
%=====================================================================
\section{Stability Analysis}
\subsection{Eigenvalue Analysis}
For a linear model like this one, whether the system is stable comes down
entirely to the eigenvalues of $A$. Each eigenvalue $\lambda$ corresponds
to a mode of motion that evolves as
\begin{equation}
x(t) \;=\; x_0\, e^{\lambda t},
\qquad \lambda = \sigma + j\omega,
\label{eq:modeevolution}
\end{equation}
so its real part $\sigma$ tells us whether that mode grows or dies out
over time, and its imaginary part $\omega$ tells us whether it oscillates
while doing so. A negative real part means the mode decays away, a
positive one means it grows without bound, and because $A$ depends on
forward speed $V$, the eigenvalues shift as speed changes. The critical
speed is simply the point where one of them crosses over from negative to
positive.
There are two different ways an eigenvalue can cross zero, and it turns
out our system does both, depending on how the trailer is loaded.
\subsection{Real Eigenvalue Crossing: Divergent Instability}
The first way is for a purely real eigenvalue to cross zero. Physically
this does not produce any oscillation at all. Once it is triggered, the
articulation angle and the lateral offset just grow steadily in one
direction, which is essentially the jackknife mode. This is what happens
in our baseline loading configuration, where the trailer's center of
gravity sits ahead of the trailer axle. Scanning forward speed from 1 to
15~m/s, the real mode stays negative, and therefore stable, until
7.5~m/s, after which it crosses zero and the system becomes divergently
unstable. Over that same speed range the oscillatory sway mode never
crosses zero at all in this configuration, so the baseline loading only
ever fails in the divergent way.
\begin{figure}[!t]
\centering
\includegraphics[width=\columnwidth]{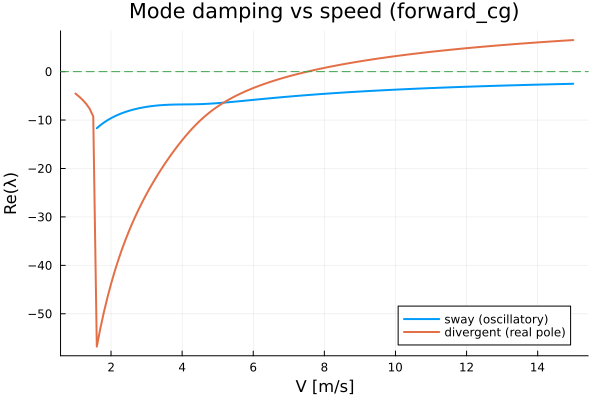}
\caption{Real part of the sway and divergent modes against forward speed for the baseline (forward-CG) loading. Only the real mode crosses zero, at 7.5~m/s; the oscillatory mode stays damped across the whole range. This is open loop.}
\label{fig:damping_fwd}
\end{figure}
\subsection{Complex Pair Crossing: Oscillatory Instability}
The second way is for a complex conjugate pair of eigenvalues to cross
the imaginary axis together. Here the motion does oscillate, at a
frequency set by the imaginary part, while the amplitude of that
oscillation grows. This is trailer sway in the proper sense of the word.
This is what happens once we move the trailer's cargo rearward, into our
destabilized configuration. The behavior is essentially the mirror image
of the baseline case. The real mode now stays damped across the whole 1
to 15~m/s range, while the oscillatory pair crosses zero at 5.4~m/s,
close to but above the 7.5~m/s where the baseline configuration failed.
\begin{figure}[!t]
\centering
\includegraphics[width=\columnwidth]{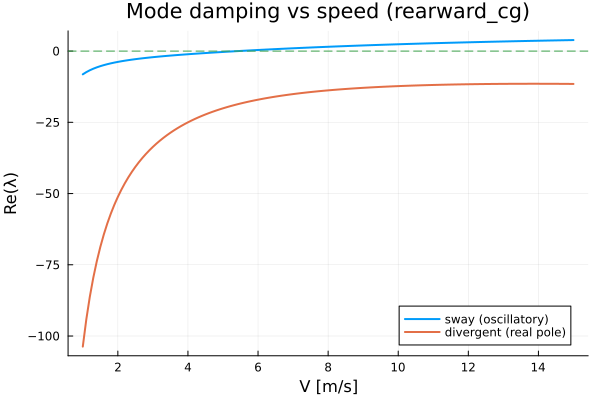}
\caption{Real part of the sway and divergent modes against forward speed for the destabilized (rearward-CG) loading. The oscillatory pair crosses zero at 5.4~m/s while the real mode stays damped, the mirror image of the baseline case. This is open loop.}
\label{fig:damping_rear}
\end{figure}
The more important takeaway here is that loading does not just move the
critical speed up or down, it changes which kind of instability actually
shows up. Moving the cargo back got rid of the divergent instability
entirely, and replaced it with an oscillatory one at a comparable speed.
That means a controller built to only catch one of these failure modes
would not necessarily catch the other, which is something we kept in mind
going into the controller design.
The two damping-vs-speed plots above each track one configuration at a
time, but it's worth seeing both crossings on the same picture. The root
locus below plots all four eigenvalues in the complex plane as speed
sweeps upward, with the baseline and destabilized configurations
overlaid, so the two different crossings are visible together rather than
in two separate plots.
\begin{figure}[!t]
\centering
\includegraphics[width=\columnwidth]{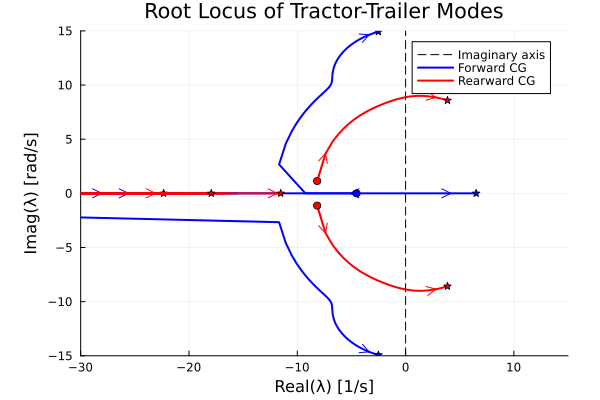}
\caption{Root locus of the four open-loop modes as forward speed sweeps from 1 to 15~m/s, with the baseline (blue) and destabilized (red) loadings overlaid. Circles mark $V_{\min}$, stars mark $V_{\max}$. The baseline crosses the imaginary axis along the real axis; the destabilized loading crosses as a complex pair.}
\label{fig:rootlocus}
\end{figure}
We also swept the trailer CG position directly to see how sharply the
stability boundary depends on loading. This sweep is run against the
reference parameter set defined at the top of the model file rather than
against the Neoracer rig, so its vertical axis should be read as a trend
rather than as critical speeds for our own platform: moving the CG
rearward compresses the stable speed range steeply at first and then more
gradually, and the steepest part of that curve is exactly where a user
loading a trailer by eye is most likely to be operating.
\begin{figure}[!t]
\centering
\includegraphics[width=\columnwidth]{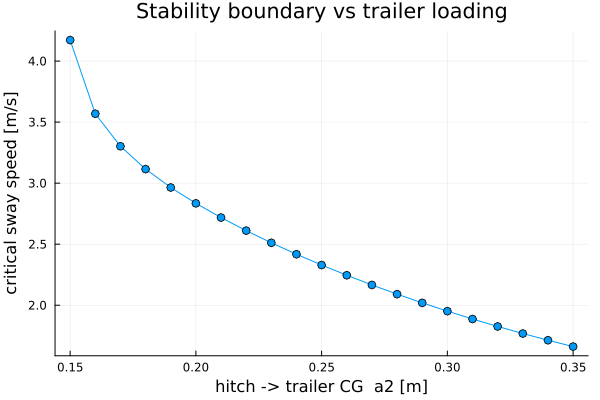}
\caption{Critical sway speed as a function of the hitch-to-trailer-CG distance $a_2$, computed against the model file's reference parameter set rather than the Neoracer rig. The trend, not the absolute speeds, is the result here: moving the CG rearward collapses the stable speed range.}
\label{fig:tongue}
\end{figure}
\subsection{Open-Loop Time-Domain Outcomes}
Eigenvalues tell us whether the system is stable, but they do not show us
what that instability actually looks like over time, so we ran
time-domain simulations to see it directly. We excited the model with a
single steering pulse of 0.05~rad, about 2.9~degrees, held from
$t = 0.5$~s to $t = 1.0$~s and then released back to zero. We ran this at
two speeds that bracket the 5.4~m/s critical speed of the destabilized
configuration: 6.6~m/s, which is 80 percent of critical, and 8.6~m/s,
which is 105 percent of critical. Picking the speeds relative to the
critical speed rather than as round numbers meant the two runs could be
compared directly against each other.
Below the critical speed, the articulation angle rings up to about 25
degrees after the pulse and then decays smoothly back toward zero within
roughly 8 seconds, which is exactly what we would expect from a stable,
damped mode.
\begin{figure}[!t]
\centering
\includegraphics[width=\columnwidth]{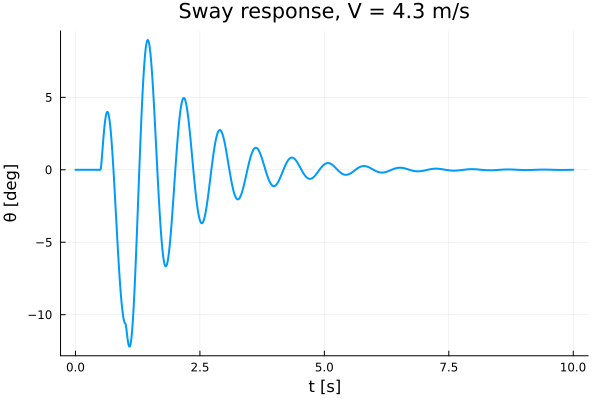}
\caption{Articulation angle response to a 0.05~rad steering pulse at 6.6~m/s, 80\% of critical. The mode rings up to about 25 degrees and decays to near zero within roughly 8~s.}
\label{fig:sway_slow}
\end{figure}
Just above the critical speed, the same pulse produces an oscillation
that grows on every cycle instead of shrinking, and it is already past
120 degrees within the 10 second simulation window.
\begin{figure}[!t]
\centering
\includegraphics[width=\columnwidth]{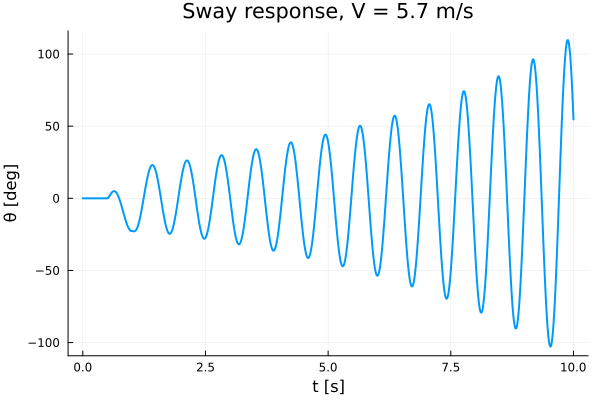}
\caption{Articulation angle response to the same 0.05~rad pulse at 8.6~m/s, 105\% of critical. The oscillation grows on every cycle and passes 120 degrees within the 10~s window.}
\label{fig:sway_fast}
\end{figure}
The difference between these two runs comes entirely from the change in
forward speed, since everything else about the pulse and the initial
conditions is identical, which demonstrates such a system's sensitivity to speed.
These simulations also gave us an independent check on the eigenvalue
analysis, since the two calculations arrive at essentially the same
answer through completely different means. At 8.6~m/s the unstable pair
sits at $0.139 \pm 9.339j$, which predicts an oscillation period of
$2\pi/9.339 = 0.673$~s. That lines up well with the peak-to-peak spacing
we measured directly off the simulated trace. It also predicts an
amplitude doubling time of $\ln(2)/0.139 = 5.0$~s, which matches how
quickly the envelope grows in the plot, roughly a factor of two between
the 2.5~s and 7.5~s peaks. At 6.6~m/s the pair sits at
$-0.662 \pm 9.432j$ instead, predicting a period of 0.666~s and an
amplitude half-life of 1.05~s, both of which line up with the decaying
run as well. Because the eigenvalues and the simulated trajectories agree
with each other on both frequency and growth or decay rate, we are
confident the modes are being computed and interpreted correctly, and
that the eigenvalue analysis from the previous section is actually
describing the system's real behavior and not just an artifact of the
linear algebra.
It is also worth looking at what happens to the vehicle's actual position
during these two runs, not just the articulation angle. Even in the
6.6~m/s case, where the articulation angle decays back to zero and the
trailer settles down, the vehicle still ends up roughly 35 meters off its
original line by the end of the run, having travelled about 66 meters
forward.
\begin{figure}[!t]
\centering
\includegraphics[width=\columnwidth]{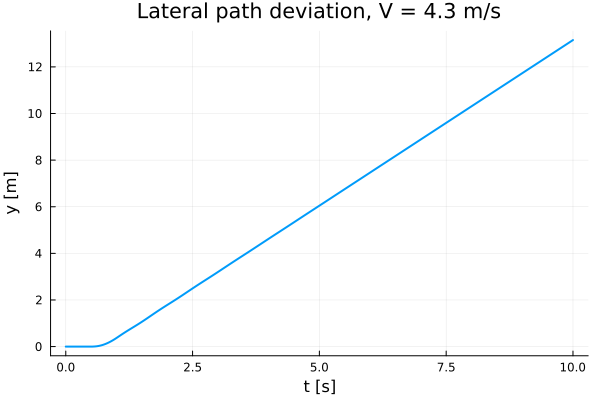}
\caption{Lateral position of the tractor against time at 6.6~m/s. Even though the articulation angle decays, the vehicle ends about 35~m off its original line.}
\label{fig:path_slow}
\end{figure}
At 8.6~m/s the drift is larger still, over 60 meters by the end of the
run, and it has the growing sway oscillation superimposed on top of it.
\begin{figure}[!t]
\centering
\includegraphics[width=\columnwidth]{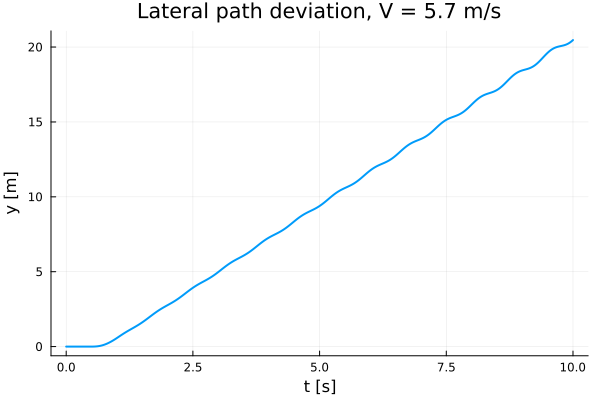}
\caption{Lateral position of the tractor against time at 8.6~m/s. The drift exceeds 60~m and the growing sway oscillation is superimposed on top of it.}
\label{fig:path_fast}
\end{figure}
Both of these happen because the steering pulse leaves the tractor with a
permanent change in heading, and nothing in the open-loop system corrects
for it, so the vehicle just keeps driving straight along its new heading.
This told us early on that the four-state model, on its own, was not
going to be enough to actually keep the vehicle in its lane, which is
what led us to add the two additional states described in the next
section.
%=====================================================================
\section{Extension to a Six-State Model}
\subsection{Why the Extension Is Needed}
While a four-state model allowed for the confirmation of trailer sway as
a problem, it was missing key components for stabilizing it. When
attempting our proposed solution on the four-state model, the car would
begin to minimize the trailer sway; however, it would completely change
its heading and drive in a different direction.
To combat this issue, two additional states that did not affect sway
dynamics themselves, but were necessary for the controller, must be
added:
\begin{itemize}
  \item $\psi_1$, the tractor's heading angle relative to the lane
        centerline;
  \item $y$, the tractor CG's lateral offset from the lane centerline.
\end{itemize}
\subsection{Derivation}
Using the small-angle assumption the new states are simple kinematic
integrators. The tractor heading $\psi_1$ integrates the tractor yaw
rate, and the lateral position $y$ integrates the total lateral velocity
in the global frame, which for small angles is the body-frame lateral
velocity plus the forward speed multiplied by the heading:
\begin{equation}
\dot{\psi}_1 = r_1,
\qquad
\dot{y} = v_1 + V \psi_1
\label{eq:newstates}
\end{equation}
Neither equation affects or introduces a new force on the rest of the
system so the rest of the matrix remains unchanged. The new system is
\begin{equation}
x = [v_1,\; r_1,\; r_2,\; \theta,\; \psi_1,\; y]^\mathsf{T},
\qquad
\dot{x} = A_6 x + B_6 \delta
\end{equation}
with
\begin{equation}
A_6 =
\left[
\begin{array}{c|c}
A_4 & \mathbf{0}_{4\times 2} \\[4pt]
\hline
\rule{0pt}{16pt}
\begin{matrix} 0 & 1 & 0 & 0 \\ 1 & 0 & 0 & 0 \end{matrix}
&
\begin{matrix} 0 & 0 \\ V & 0 \end{matrix}
\end{array}
\right],
\qquad
B_6 =
\begin{bmatrix} B_4 \\ 0 \\ 0 \end{bmatrix}
\label{eq:A6}
\end{equation}
The new matrix has $A_4$, the original state matrix, in its upper block,
with the upper-right block of the matrix that corresponds to the new
states' effect on the original states set to zero, since the original
states feed into the new states, but the new states do not feed back into
them. The final two rows of the matrix show the simple kinematic
integrator equations. Therefore, the eigenvalues of the four-state model
remain eigenvalues of the six-state model. The extension adds two
integrator poles at the origin and does not change any of the stability
conclusions from Section~IV.
%=====================================================================
\section{LQR Controller Design}
\subsection{What Is LQR}
A linear quadratic regulator computes a state feedback control law that
minimizes a quadratic cost accumulated over infinite time:
\begin{equation}
J = \int_{0}^{\infty}
\left( x^\mathsf{T} Q\, x + u^\mathsf{T} R\, u \right) dt
\label{eq:lqrcost}
\end{equation}
The matrix $Q$ assigns a penalty to each state deviation and $R$ assigns
a penalty to control effort. Therefore minimizing $J$ gives the feedback
law
\begin{equation}
u = -K x,
\qquad
K = R^{-1} B^\mathsf{T} P
\label{eq:lqrlaw}
\end{equation}
where $P$ is the positive definite solution of the algebraic Riccati
equation
\begin{equation}
A^\mathsf{T} P + P A - P B R^{-1} B^\mathsf{T} P + Q = 0
\label{eq:riccati}
\end{equation}
We chose LQR for three reasons. First, it handles multi-objective
trade-offs directly, so sway damping, lane keeping, and steering economy
become three entries in the $Q$ and $R$ matrices rather than three
separate control loops that have to be reconciled with each other.
Second, it is a full-state feedback method, and our six-state model is
small enough that full-state feedback is realistic. Third, the resulting
closed-loop system has guaranteed stability margins as long as the plant
is controllable, which matters here because the open-loop plant is
unstable by construction.
\subsection{Our Controller Implementation}
We compared candidate weightings on three criteria: how quickly the
articulation angle settles, how far the vehicle drifts laterally, and how
much steering angle the controller demands. We chose to penalize outcomes
rather than internal states. The final weighting places cost only on the
three states that describe whether the vehicle is actually failing,
articulation angle, heading error, and lateral position error, and
places no cost at all on tractor lateral velocity or on either yaw rate.
Those three unpenalized states still show up in the feedback law, but
only because they are dynamically coupled to the penalized ones, not
because we asked the controller to regulate them directly:
\begin{equation}
Q = \mathrm{diag}\!\left(0,\; 0,\; 0,\; 50,\; 20,\; 25\right),
\qquad
R = 100
\label{eq:QR}
\end{equation}
\begin{table}[t]
\caption{LQR weighting rationale. The state order is
$[v_1, r_1, r_2, \theta, \psi_1, y]$.}
\label{tab:qr}
\centering
\footnotesize
\begin{tabular}{@{}llp{3.6cm}@{}}
\toprule
State & Weight & Why \\
\midrule
$v_1$    & $0$   & internal state, not a failure mode; regulated only through coupling \\
$r_1$    & $0$   & internal state, not a failure mode \\
$r_2$    & $0$   & internal state, not a failure mode \\
$\theta$ & $50$  & largest weight: sway is the failure mode we are targeting \\
$\psi_1$ & $20$  & heading error; with $y$ this makes lane keeping the second objective \\
$y$      & $25$  & lane position error \\
\midrule
$\delta$ & $R=100$ & larger than any state weight, to keep steering demand small \\
\bottomrule
\end{tabular}
\end{table}
We penalized only the three states that describe failure: articulation
angle, heading error, and lateral position error. Tractor lateral
velocity and both yaw rates have to change during any correction, and a
penalty on them would make the controller resist the motion it needs to
bring the trailer back in line, so their weights are zero.

Articulation angle carries the largest weight, 50, because sway is the
failure mode we are targeting. Heading and lane position together carry
45, which keeps lane keeping nearly as important as sway damping.

\begin{figure}[!t]
\centering
\includegraphics[width=\columnwidth]{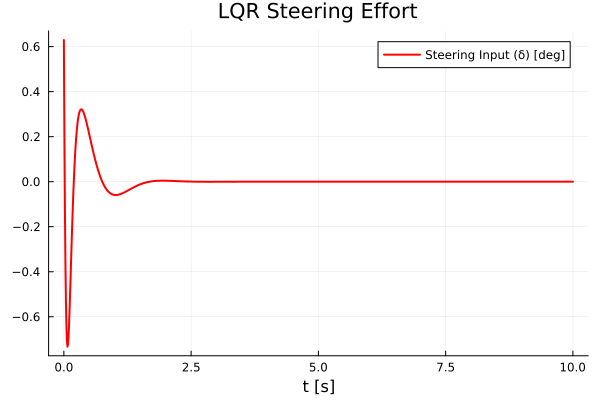}
\caption{Commanded steering angle against time under full-state LQR feedback at 8.6~m/s. Peak demand is about 0.77 degrees and the input decays to zero within about 1.5~s.}
\label{fig:lqr_effort}
\end{figure}

Only the ratio of the state weights to $R$ affects the gain, since
scaling $Q$ and $R$ by the same constant scales $P$ by that constant and
leaves $K$ unchanged. We set $R = 100$, above every state weight, so the
controller accepts a slightly slower settling time in exchange for a much
smaller steering demand. For each candidate weighting we solved the
Riccati equation at the design speed, simulated recovery from a 5 degree
initial articulation angle, and compared articulation settling time,
lateral drift, and peak steering angle.
The final weighting settles articulation within about a second, keeps
lateral position essentially flat, and peaks at about 0.77 degrees of
steering (Fig.~\ref{fig:lqr_effort}).
\begin{equation}
K =
\setlength{\arraycolsep}{2.5pt}
\begin{bmatrix}
-0.1185 & 0.0708 & 0.0650 & -0.1258 & 1.2819 & 0.5000
\end{bmatrix}
\label{eq:K}
\end{equation}
\begin{table}[t]
\caption{Where each LQR gain acts, and whether the state it multiplies is
directly measurable on the tow vehicle. Gains solved at the $V=8.6$ m/s
design speed.}
\label{tab:gains}
\centering
\footnotesize
\begin{tabular}{@{}lllc@{}}
\toprule
State & Meaning & Gain & Measurable on tractor? \\
\midrule
$v_1$    & tractor lateral velocity & $-0.1185$ & indirect (integrated accel.) \\
$r_1$    & tractor yaw rate         & $0.0708$  & yes (yaw gyro) \\
$r_2$    & trailer yaw rate         & $0.0650$  & \textbf{no} (needs estimator) \\
$\theta$ & articulation angle       & $-0.1258$ & \textbf{no} (needs estimator) \\
$\psi_1$ & heading vs.\ lane        & $1.2819$  & yes (LiDAR wall fit) \\
$y$      & lateral offset from lane & $0.5000$  & yes (LiDAR wall fit) \\
\bottomrule
\end{tabular}
\end{table}
Two of these gains act on states the tow vehicle cannot measure. In
practice they multiply estimates supplied by the Kalman filter of
Section~\ref{sec:estimation}, and the closed-loop results there show
that this costs little in performance.

The gain on lateral position comes out to exactly 0.5000, which looks at
first like a hardcoded value but is not. For the outermost integrator in
a controllable chain, the LQR solution reduces to the square root of
$Q_{yy}$ over $R$, and the square root of $25/100$ is 0.5 exactly. We
take this as a useful sanity check that the Riccati solve is behaving as
expected. The same shortcut does not hold for the other states, since for
example the square root of $Q_{\psi\psi}$ over $R$ comes out to 0.447
while the actual heading gain is 1.2819.
\subsection{Controller Results}
\begin{figure}[!t]
\centering
\includegraphics[width=\columnwidth]{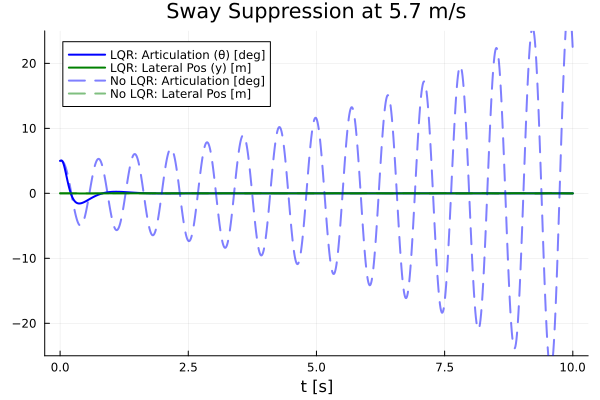}
\caption{Articulation angle and lateral position at 8.6~m/s in the destabilized configuration, with the LQR active (solid) and with no controller (dashed), both released from a 5 degree initial articulation angle. At 105 percent of this configuration's 5.4~m/s critical speed the uncontrolled articulation angle diverges to about 17 degrees by 10~s, while the controlled response settles within about a second and the lateral position stays flat.}
\label{fig:lqr_states}
\end{figure}

Figure~\ref{fig:lqr_states} shows the closed-loop response at 8.6~m/s in
the destabilized configuration, the same 105 percent of critical speed
used for the open-loop runs in Section~IV. At this speed the open-loop
sway pair sits at $+0.139 \pm 9.339j$, so the uncontrolled system is
genuinely unstable rather than merely lightly damped, and the run uses
the design-point gain of \eqref{eq:K} directly. Unlike the open-loop runs in
Section~IV, this simulation is released from an initial articulation
angle of 5 degrees with no steering pulse, representing a trailer that
has already been knocked out of line and needs to be recovered. We
overlaid the uncontrolled response from the same initial condition for
comparison.
Because 8.6~m/s sits above the 5.4~m/s critical speed of this loading,
the comparison is a stable-versus-unstable one. The uncontrolled trace
never recovers: the 5 degree initial articulation angle rings up on every
cycle, passing 10 degrees at about 6~s and reaching roughly 17 degrees by
the end of the 10~second window, an envelope growing at the 0.14 per
second set by the unstable pole. The controlled trace decays to zero
within about a second. What the run establishes is that a single fixed
gain, solved at this speed, both stabilizes a sway mode that is diverging
on its own and closes the lane-keeping loop the open-loop system does not
have at all, without demanding meaningful steering authority. The MuJoCo
results in Section~IX provide the same comparison in a nonlinear
multibody simulation.
The controlled articulation angle overshoots once to about $-1.5$
degrees at 0.3~s, is back inside half a degree by 0.6~s and has settled
by roughly 1~s, while the lateral position trace stays essentially flat
for the full 10~s. Most importantly, the controller does
this without trading sway damping against lane position, since both
objectives are met at the same time. That was the requirement that ruled
out the human-driver-in-the-loop assumption made by prior work.
The control effort is small in both magnitude and duration. Peak steer
demand is about 0.77 degrees, well under the one degree of authority the
weighting was chosen to respect, and the input decays to zero within
1.5~s, meaning the intervention would be imperceptible to a passenger and
well within the rate and travel limits of a production steering
actuator.
This matters for the practical argument of the paper, because the
alternative approaches we surveyed in Section~II require hardware the
vehicle does not have, whereas ours requires actuator authority the
vehicle already has in abundance.
The same result can be stated in terms of eigenvalues, which connects
this section back to the stability argument of Section~IV.
\begin{table}[t]
\caption{Open-loop vs.\ closed-loop poles of the six-state destabilized
configuration at the $V = 8.6$ m/s design speed. Modes are paired by
continuity along $A - tBK$ as $t$ runs from 0 to 1.}
\label{tab:poles}
\centering
\footnotesize
\begin{tabular}{@{}lll@{}}
\toprule
Mode & Open loop & Closed loop ($u = -Kx$) \\
\midrule
lane integrators & $0,\; 0$                 & $-2.794 \pm 3.678j$ \\
sway pair        & $+0.139 \pm 9.339j$      & $-19.068 \pm 17.067j$ \\
fast real \#1    & $-9.307$                 & $-9.824$ \\
fast real \#2    & $-28.626$                & $-22.403$ \\
\bottomrule
\end{tabular}
\end{table}
The controller does three things at once here. It takes the unstable
sway pair from $+0.139 \pm 9.339j$ deep into the left half plane at
$-19.068 \pm 17.067j$, it pulls the two integrators at the origin away
from the imaginary axis, which is what closes the lane keeping loop, and
it leaves the two real modes roughly where they were, since those were
already well damped and there is nothing to gain from moving them.
Worth noting is the reversal against the open loop. Once the loop is
closed the sway mode is no longer the limiting one, because the
controller has moved it further than any other mode in the system. The
slowest closed-loop pole is now the lane-keeping pair at
$-2.794 \pm 3.678j$, a time constant of 0.36~s and therefore a settling
time of a little over one second, consistent with the roughly one second
settling observed in Fig.~\ref{fig:lqr_states}. What limits the
closed-loop response at the design speed is lane keeping, not sway.
One consequence of designing at a single speed is worth stating, because
it decides how the controller would be deployed. The gain vector is
solved from the $A$ matrix at one value of $V$, and $A$ depends on $V$
throughout, so nothing about the design guarantees the gain away from its
design point. It has to be checked, and the direction of the check
matters.
Designing at the top of the range holds. Applying the 8.6~m/s gain of
\eqref{eq:K} unchanged at 6.6~m/s leaves the worst closed-loop pole at
$-3.157$, and at 3.0~m/s at $-0.923$; both are stable, though the margin
does thin out at the low end as that slowest pole drifts toward the
imaginary axis. Designing at the bottom does not hold. A gain solved
instead at 3.0~m/s and applied at 8.6~m/s leaves the closed-loop sway
pair at $+0.163 \pm 6.065j$, which is unstable, so a controller tuned on
the safe part of the envelope fails on exactly the part it is meant to
protect.
That asymmetry is the practical result. A single fixed gain is enough
here provided it is solved at the fastest speed the vehicle is expected
to tow at, which is why we design at 8.6~m/s rather than in the middle of
the range. A deployed version could still schedule the gain on measured
forward speed, which the vehicle already has from wheel encoders, and
would want to if the operating range were wider than the one we simulate,
but the results here do not require it.
%=====================================================================
\section{Sensing and State Estimation}
\label{sec:estimation}
\subsection{Sensor Feasibility on the Vehicle}
As set out in the scope of Section~II, the trailer carries no sensors,
while the Neoracer, like most autonomous vehicles, can already measure
its own motion. The controller still needs all six states, including the
two that belong to the trailer, trailer yaw rate and articulation angle,
and neither can be measured without instrumenting the trailer.

We first tested the LQR using only the four states available from the
vehicle. This four-state configuration could not stabilize the system
without substantial tuning per car setup. We therefore designed a Kalman
filter to estimate the remaining two states from the four values we could
measure.
\subsection{Kalman Filter Implementation in Julia}
Our four measurable states, tractor lateral velocity, tractor yaw
rate, tractor heading, and lateral position, are extracted through a
measurement matrix $C$ that picks out those rows of the six-state vector
and discards the other two. This gives us a standard linear measurement
model
\begin{equation}
y = C x,
\qquad
C =
\begin{bmatrix}
1 & 0 & 0 & 0 & 0 & 0 \\
0 & 1 & 0 & 0 & 0 & 0 \\
0 & 0 & 0 & 0 & 1 & 0 \\
0 & 0 & 0 & 0 & 0 & 1
\end{bmatrix}
\label{eq:C}
\end{equation}
where the trailer yaw rate and articulation angle are simply absent from
what we observe directly.
Rather than hand-picking the process and measurement noise covariances,
$Q_{kf}$ and $R_{kf}$, we tuned them numerically. We treated their
diagonal entries as free parameters (in log space, to keep them positive
and to search efficiently across several orders of magnitude) and
optimized them to minimize the squared error between the true and
estimated values of the two unmeasured states, trailer yaw rate and
articulation angle, over a short simulation window. We included the
lateral position estimate in this cost as a sanity check, since it is
directly measured and should be estimated almost perfectly regardless of
tuning. The cost function minimized was
\begin{equation}
\mathcal{J}_{est} = \sum_{i}
5 \left(\theta_i - \hat{\theta}_i\right)^2 +
5 \left(r_{2,i} - \hat{r}_{2,i}\right)^2 +
\left(y_i - \hat{y}_i\right)^2
\label{eq:estcost}
\end{equation}
For each candidate set of covariances, we built the corresponding
steady-state Kalman gain $L$ using Julia's built-in \texttt{kalman}
function, then simulated the closed-loop LQG system (LQR feedback acting
on the estimated state rather than the true state) and scored how closely
the estimator tracked the true states. We used Nelder--Mead optimization,
bounded in log space, to search over this ten-dimensional parameter space
(six process noise terms and four measurement noise terms).
\begin{figure}[!tb]
\centering
\includegraphics[width=\columnwidth]{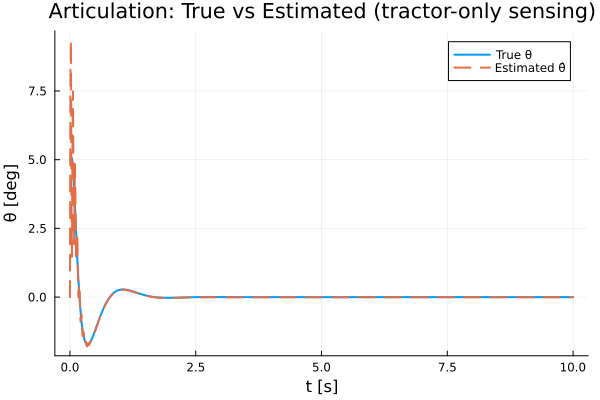}
\caption{Articulation angle, true against Kalman-estimated, under tractor-only sensing at 8.6~m/s. The estimate overshoots to about 7.9 degrees during the first 0.15~s while the filter has no information, then converges.}
\label{fig:lqg_theta}
\end{figure}
\begin{figure}[!tb]
\centering
\includegraphics[width=\columnwidth]{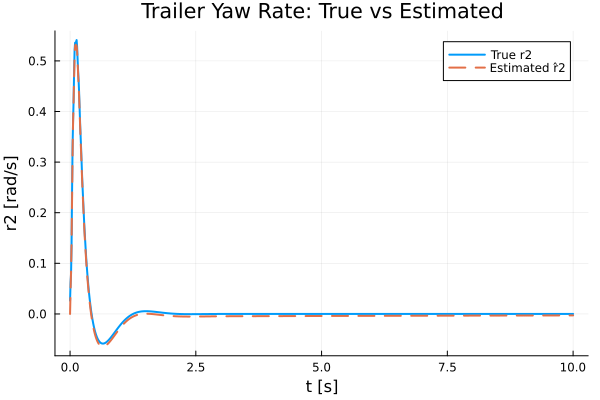}
\caption{Trailer yaw rate, true against Kalman-estimated, under tractor-only sensing at 8.6~m/s. The estimate tracks the 0.68~rad/s peak to within about two percent and lags slightly on the return.}
\label{fig:lqg_r2}
\end{figure}
Initially, letting the optimizer search freely produced an estimator that
tracked trailer yaw rate reasonably well but left the articulation angle
estimate noticeably lagging and slow to settle. The optimizer was driving
the process noise for $\theta$ up against the edge of our search space,
effectively telling the filter to distrust its own model of articulation
dynamics and lean almost entirely on the tractor-only measurements to
reconstruct it, which is a difficult thing to do well since $\theta$ has
no direct measurement path at all. We bounded the log-space search to
keep this term, and the rest of the covariances, within a range that
still let the optimizer find a well-conditioned solution rather than
drifting toward degenerate extremes. Even with that constraint in place,
the process noise on $\theta$ still settled near the upper edge of the
bound at 122.2, meaning the filter genuinely does need to lean hard on
inference to recover this state, but the bound kept it from destabilizing
the rest of the estimate in the process. One other term finished against
the opposite bound, the measurement noise on $y$, so two of the ten tuned
entries are set by the bound rather than by the data. Nelder--Mead is also a local search, so what we
report is one converged solution rather than a global optimum. Both are
reasons to treat the covariances as a working tuning rather than an
identified noise model.
\begin{table}[t]
\caption{Final tuned noise covariances from the bounded Nelder--Mead
search at the $V = 8.6$ m/s design speed. Search bounds were $e^{-10}$ to
$e^{5}$ on every diagonal entry. Entries marked $\dagger$ finished
against a search bound.}
\label{tab:cov}
\centering
\footnotesize
\begin{tabular}{@{}llll@{}}
\toprule
\multicolumn{2}{@{}l}{Process noise $W$} &
\multicolumn{2}{l}{Measurement noise $V_n$} \\
\cmidrule(r){1-2}\cmidrule(l){3-4}
State & Value & Measurement & Value \\
\midrule
$v_1$    & $4.32 \times 10^{-4}$ & $v_1$    & $5.74 \times 10^{-3}$ \\
$r_1$    & $8.0 \times 10^{-5}$  & $r_1$    & $4.29 \times 10^{-4}$ \\
$r_2$    & $8.9 \times 10^{-5}$  & $\psi_1$ & $8.54 \times 10^{-4}$ \\
$\theta$ & $48.26$\,$^\dagger$   & $y$      & $1.17 \times 10^{-3}$\,$^\dagger$ \\
$\psi_1$ & $6.0 \times 10^{-5}$  &          &          \\
$y$      & $9.09 \times 10^{-4}$ &          &          \\
\bottomrule
\end{tabular}
\end{table}
\begin{figure*}[!tb]
\centering
\includegraphics[width=0.74\textwidth]{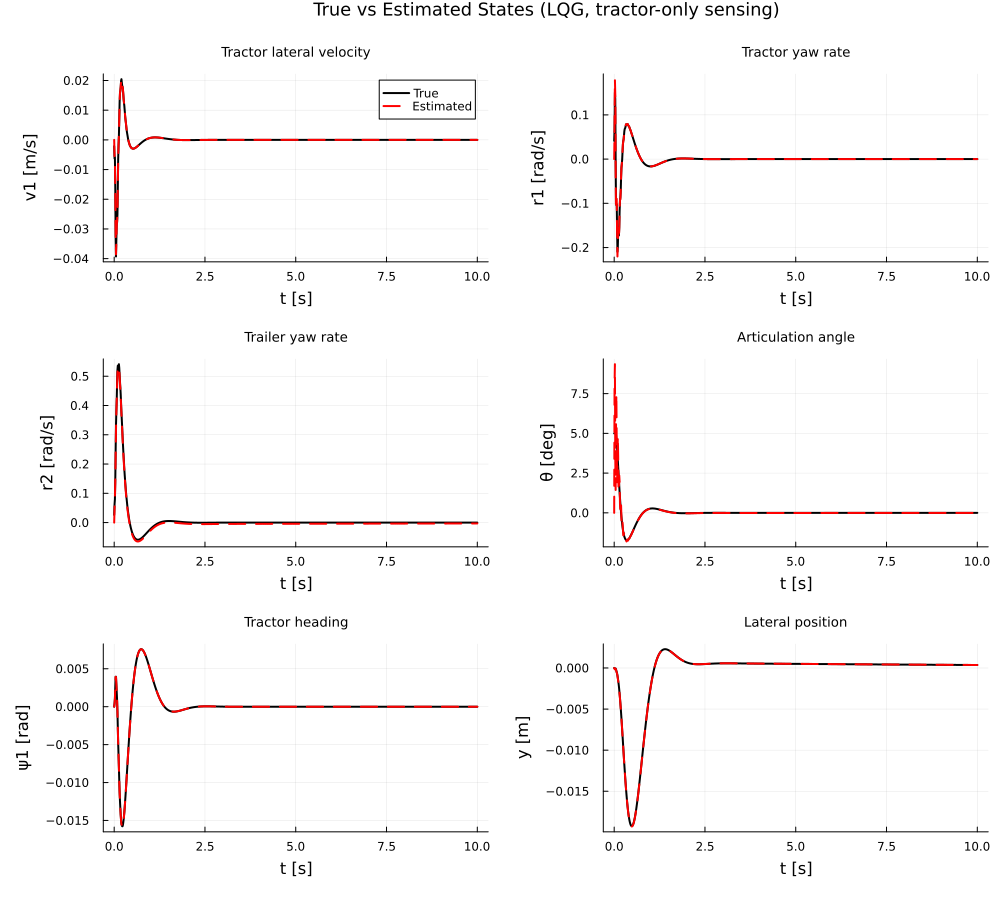}
\caption{All six states, true (black) against estimated (red dashed),
under LQG with tractor-only sensing at 8.6~m/s. The top row and bottom
row are directly measured; the middle row, trailer yaw rate and
articulation angle, is reconstructed by the filter.}
\label{fig:lqg_all}
\end{figure*}
\begin{figure*}[!tb]
\centering
\includegraphics[width=0.74\textwidth]{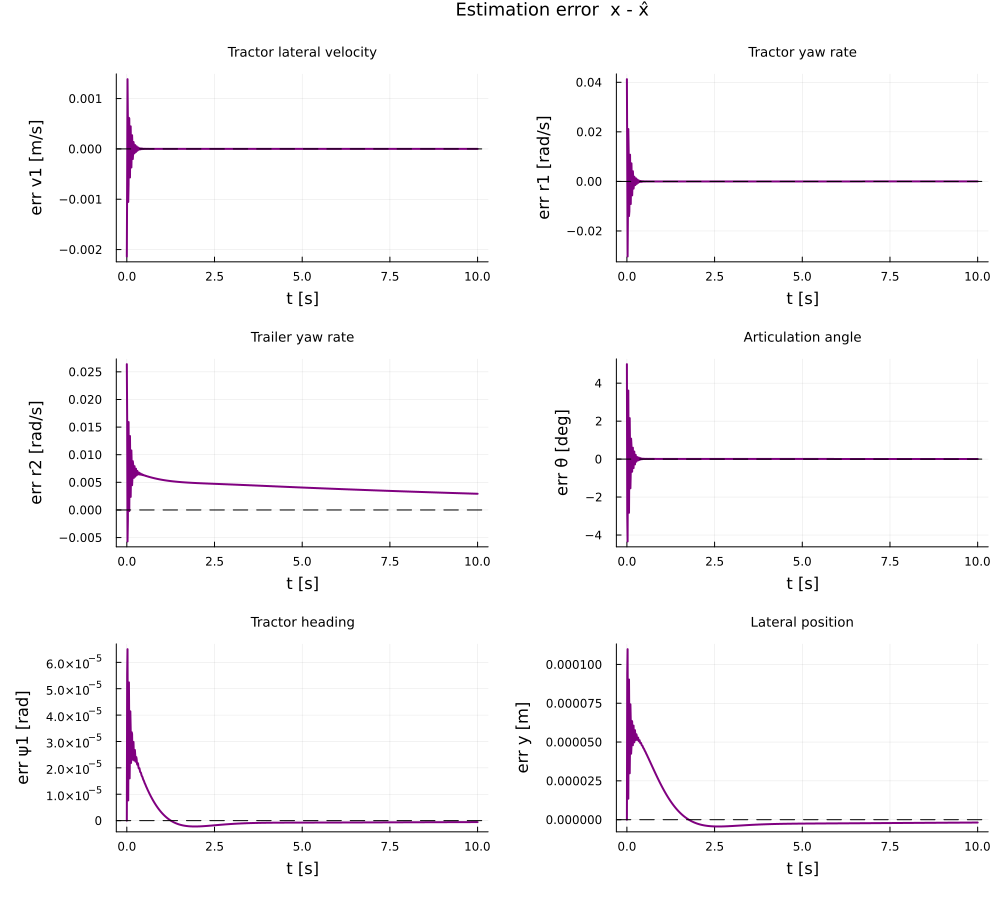}
\caption{Estimation error $x - \hat{x}$ for all six states at 8.6~m/s.
The four measured states and the articulation angle collapse within about
a second; the trailer yaw-rate error carries a long tail that is still
visible at 10~s.}
\label{fig:lqg_err}
\end{figure*}
The final tuned covariances gave process noise values of
$4.32\times10^{-4}$ ($v_1$), $8.0\times10^{-5}$ ($r_1$),
$8.9\times10^{-5}$ ($r_2$), $48.26$ ($\theta$), $6.0\times10^{-5}$
($\psi_1$), and $9.09\times10^{-4}$ ($y$), with measurement noise of
$5.74\times10^{-3}$ ($v_1$), $4.29\times10^{-4}$ ($r_1$),
$8.54\times10^{-4}$ ($\psi_1$), and $1.17\times10^{-3}$ ($y$). We
confirmed the resulting estimator error dynamics were stable, with all
six eigenvalues of $A - LC$ carrying negative real parts at the 8.6~m/s
design speed ($-90.9 \pm 87.1j$, $-19.3$, $-1.5 \pm 1.2j$ and $-0.065$),
and the optimization converged to a final cost of $0.210$. With a stable, tuned estimator in
hand, the estimator itself evolves according to
\begin{equation}
\dot{\hat{x}} = A \hat{x} + B u + L \left( y - C \hat{x} \right)
\label{eq:kfupdate}
\end{equation}
where the steady-state gain $L$ solves the dual Riccati equation
\begin{equation}
A P + P A^\mathsf{T} - P C^\mathsf{T} V_n^{-1} C P + W = 0,
\qquad
L = P C^\mathsf{T} V_n^{-1}
\label{eq:kfriccati}
\end{equation}
We simulated the full LQG system starting from the same initial
articulation angle used in our earlier full-state feedback test and at
the same 8.6~m/s, this time feeding the controller only the estimated
state vector rather than the true one.
\begin{figure}[!tb]
\centering
\includegraphics[width=\columnwidth]{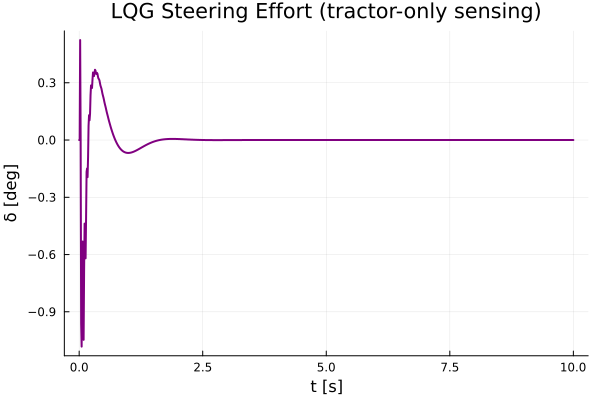}
\caption{Commanded steering angle under LQG (estimated-state) feedback at 8.6~m/s. Peak demand is about 0.96 degrees against 0.77 degrees for full-state feedback, and the initial chatter reflects the estimator transient rather than vehicle dynamics.}
\label{fig:lqg_effort}
\end{figure}
Figures \ref{fig:lqg_theta} and \ref{fig:lqg_r2} show the estimator
tracking the true articulation angle and trailer yaw rate closely after a
brief transient, confirming that
tractor-only sensing is sufficient to recover the two unmeasured trailer
states once the covariance search was properly bounded.
\begin{figure*}[!tb]
\centering
\includegraphics[width=0.80\textwidth]{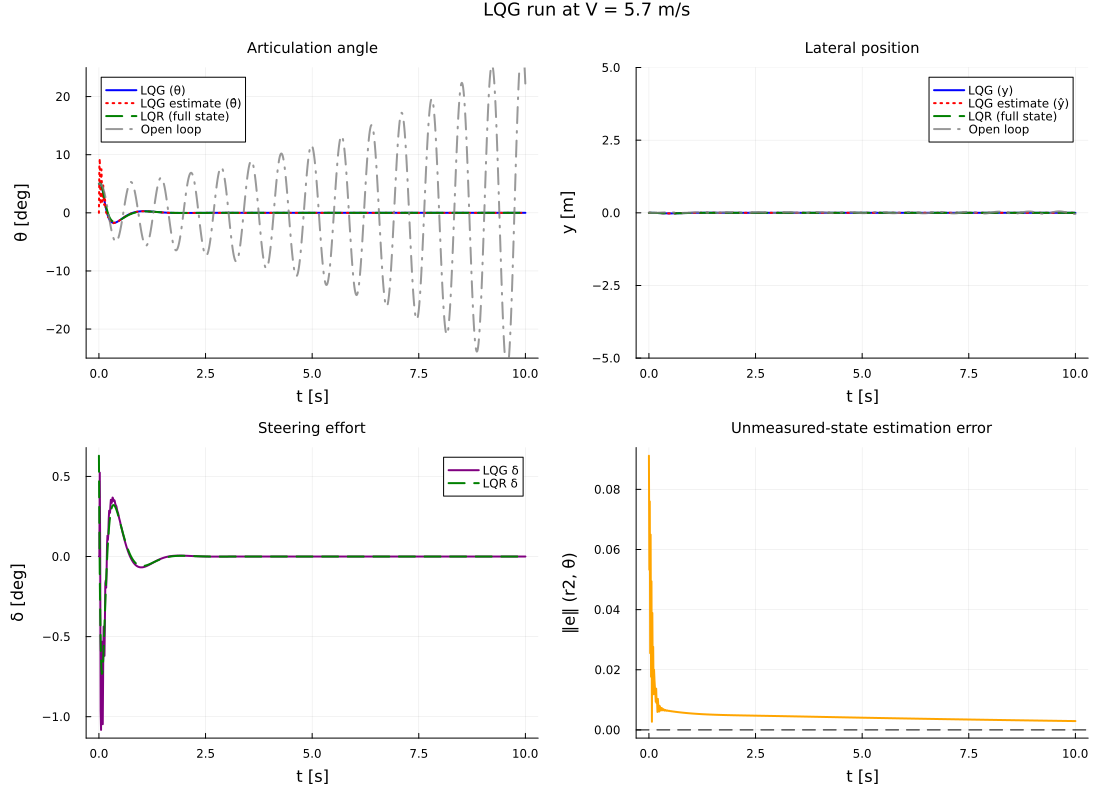}
\caption{LQG run at 8.6~m/s compared against full-state LQR and the
uncontrolled system, all released from the same 5 degree initial
articulation angle. Estimated-state feedback is nearly indistinguishable
from full-state feedback in both the response and the steering command,
while the uncontrolled articulation angle diverges past 17 degrees.}
\label{fig:lqg_run}
\end{figure*}
Figure~\ref{fig:lqg_all} extends that comparison to the whole state
vector. The four measured states are tracked essentially exactly, which
is what we would expect, and the two unmeasured trailer states are
recovered to within the transient visible in the first few tenths of a
second.
Plotting the difference rather than the two traces makes the structure of
the estimation error easier to read.
Figure~\ref{fig:lqg_err} shows two different behaviors. Five of the six
errors collapse to near zero within about a second, once the initial
transient from a zero-information starting guess has passed. The trailer
yaw-rate error behaves differently: it settles quickly to about
0.008~rad/s and then bleeds off slowly, reaching roughly 0.003~rad/s but
still not zero after 10~s. That tail is the slowest eigenvalue of
$A - LC$ at $-0.065$, a time constant of 15.4~s, and it is the honest
limit on tractor-only sensing here. It is also the one place where
designing at the top of the speed range costs something: the same
observer tuned at 3.0~m/s puts that mode at $-0.205$, so the trailer
yaw-rate estimate cleans up about twice as fast there. It is a small, slowly decaying bias rather than a growing error, so
it does not threaten stability, and because the controller weights
trailer yaw rate at zero it barely reaches the steering command at all.
It does mean the filter should not be relied on for an absolute
trailer-yaw-rate reading on a short window.
Finally, we compared the steering effort commanded under estimated-state
feedback against our earlier full-state feedback result, both at
8.6~m/s, to confirm that closing the loop through the estimator did not
come at the cost of impractical control demands.
The LQG steering input remains smooth and within a reasonable range,
peaking at about 0.96 degrees against 0.77 degrees for full-state
feedback, with the extra demand confined to the first 0.3~s while the
estimator is still converging from a zero initial guess. This confirms
that estimating the unmeasured states from tractor-only sensing does not
meaningfully degrade the control system's practicality.
Figure~\ref{fig:lqg_run} puts all three cases on the same axes at
8.6~m/s, above the critical speed: LQG on the estimated state, LQR on the
true state, and the uncontrolled system. The two controlled articulation
and lateral position traces are close enough to be hard to separate by
eye, which is the result we wanted, while the uncontrolled trace diverges
past 17 degrees over the same window. The steering commands differ only
during the estimator transient, where the LQG input peaks at about 0.96
degrees against about 0.77 degrees for full-state feedback, and both have
decayed to zero within about 1.5~s. The fourth panel tracks the norm of
the error in the two unmeasured states, and shows the same fast collapse
followed by a slow tail that Fig.~\ref{fig:lqg_err} breaks out state by
state. The norm falls from 0.107 to about 0.008 within half a second and
is still near 0.003 at 10~s.
%=====================================================================
\section{MuJoCo Simulation}
\subsection{Purpose and Assumptions}
We made two simulations of our system in MuJoCo, an open source physics
simulation software developed by Google. The goal of these simulations is
to act as a visual, more accurate representation of the state-space
models we made of the system. Additionally, MuJoCo is more accurate in
certain areas, because unlike our state-space models, we did not need to
assume that the system had a linear response to inputs, and we did not
restrict it to only one plane. We made several assumptions in our MuJoCo
model, including:
\begin{itemize}
  \item Only the truck has suspension.
  \item Tires are rigid and not compressible.
  \item Cargo is fixed onto the trailer and will not move inside the
        trailer.
  \item All six tires in the system have the same coefficients of static
        and kinetic friction.
  \item System is on a flat, straight road with consistent surface
        conditions throughout.
\end{itemize}
We made one simulation on the scale of the Neoracer vehicle, with
accurate masses and the same CAD models of the Neoracer and trailer that
we 3D printed. We made another simulation on the scale of a real-life
truck and trailer, using more generic shapes as vehicles. MuJoCo allows
us to track various metrics from the simulation; the most important
metrics we tracked are IMU measurements from the perspective of the
truck, which include linear acceleration from the accelerometer, angular
acceleration from the gyroscope, and heading from the magnetometer. From
our observations in this data, there are several characteristics of these
metrics that we found to be indicators of trailer sway occurring,
including increased angular acceleration and decreased lateral
acceleration, both due to the rear tires losing traction.
\subsection{Simulation Configuration}
The Neoracer-scale model is defined in \texttt{rc-truck-trailer.xml} and
simulated by \texttt{rc\_sway\_experiment.py}. We ran tests using two modes using simulated LiDAR data. We created a standard, PD wall-following controller to analyze the sway the system experienced without a controller designed to dampen oscillations and stabilize the system. Additionally, we implemented our LQR controller to test its efficacy in a more realistic simulation. We tested at 2.96~m/s, as we observed that the minimum speed to cause trailer sway using the PD wall follower was 2.68~m/s. Our testing setup consists of a straight track, with a 1.9~m distance between walls on either side. The trailer's cargo, a 3.5~kg square mass, was adjustable and placed 0.04~m behind the trailer's axle. The Neoracer had a mass of 2.5~kg, and the trailer itself had a mass of 0.8~kg. The simulated LiDAR specifications match that of the real Neoracer (the Richbeam LakiBeam 1L), with 1080 rays that range from -135 degrees to +135 degrees, 0 at the front of the vehicle, and scanning at 20~Hz. The LiDAR has a gaussian noise standard deviation of $\pm2$~cm. A disturbance is applied to the system by turning the wheels 5.7~degrees left for 0.4~s, then 5.7 degrees right for 0.4~s, then straight for 0.4~s.
MuJoCo's fluid simulation is used to model air resistance and aerodynamics. Additionally, this can be used to simulate a gust-type disturbance, rather than a swerve.
The commanded steering angle is passed through a slew-rate limit of
13~rad/s, representing a digital RC servo doing 60 degrees in
approximately 0.08~s.
%=====================================================================
\section{Results}
\subsection{MuJoCo Results}
\begin{figure*}[!t]
\centering
\includegraphics[width=0.72\textwidth]{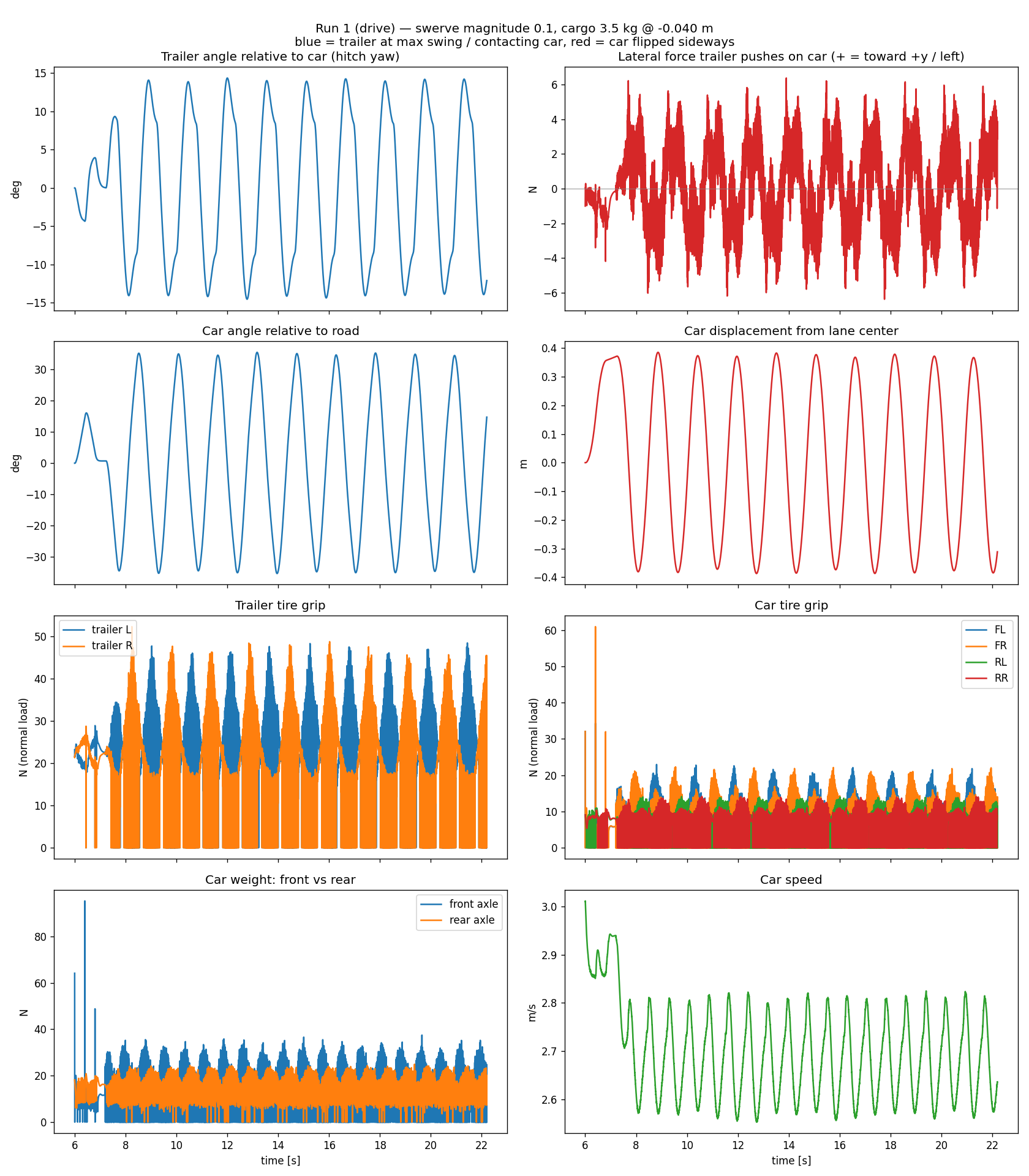}
\caption{Eight-panel MuJoCo run under PD wall-following control at
2.96~m/s with cargo 3.5~kg placed 0.040~m behind the trailer axle,
following a 0.1~rad swerve disturbance. The hitch oscillation does not
decay: it grows out of the disturbance and holds a limit cycle of about
$\pm 12$ degrees for the rest of the run, with the tow vehicle swinging
$\pm 35$ degrees against the road and crossing the full width of the
lane on every cycle.}
\label{fig:mujoco_pd}
\end{figure*}

\begin{figure*}[!t]
\centering
\includegraphics[width=0.72\textwidth]{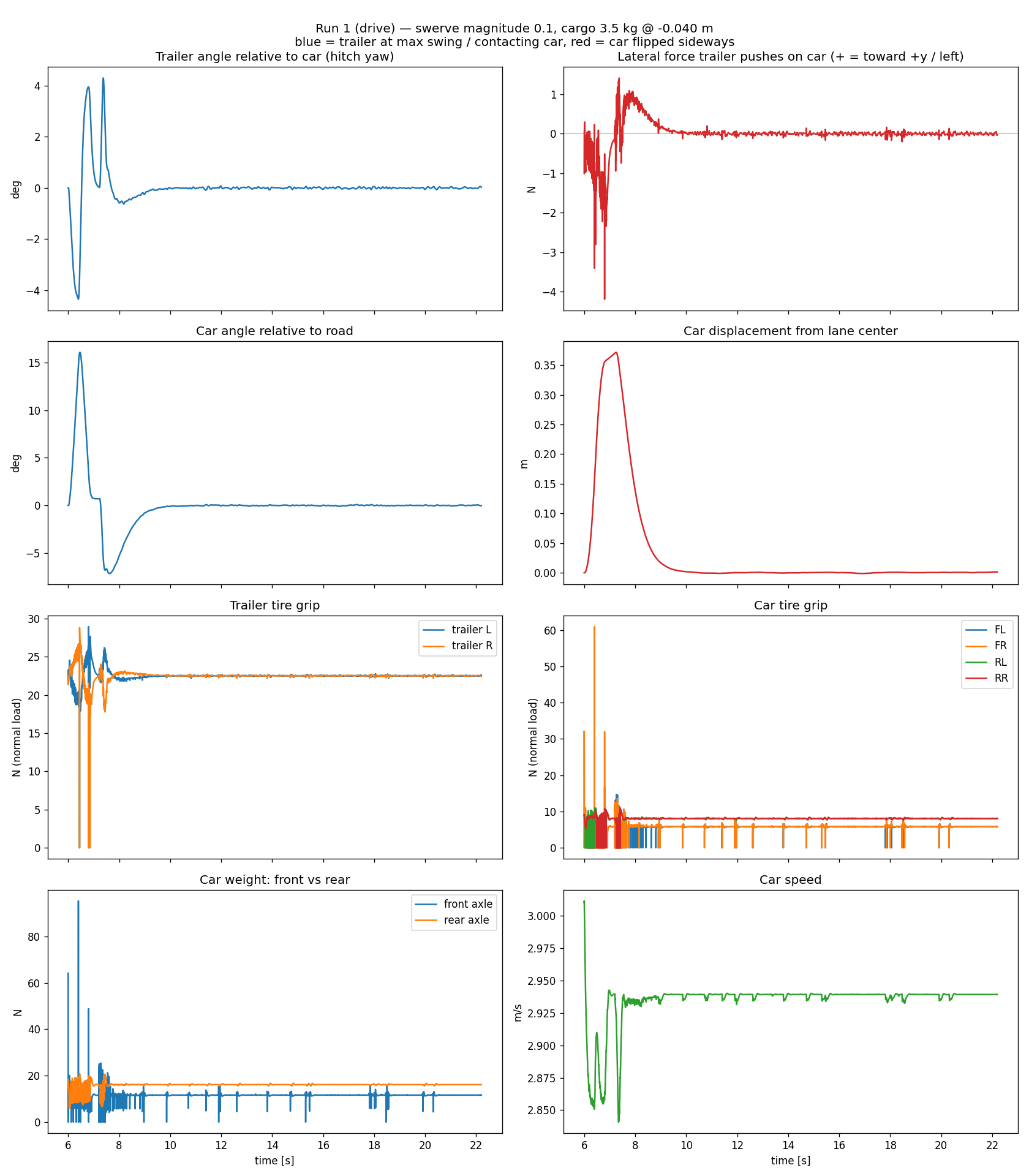}
\caption{Eight-panel MuJoCo run under six-state LQR control at 2.96~m/s with cargo 3.5~kg placed 0.040~m behind the trailer axle, following a 0.1~rad swerve disturbance. Peak hitch yaw is 4.3 degrees and both the hitch angle and the lane offset return to zero within about 2~s of the disturbance ending.}
\label{fig:mujoco_lqr}
\end{figure*}

Our testing observed that trailer sway reliably occurred for speeds of
2.68~m/s or higher when using a standard PID wall follower, without sway
prevention. At our testing speed of 2.96~m/s, over four runs, we observed
a mean of means sway amplitude of 12.505 degrees with a standard
deviation of 0.015 degrees. We observed a mean max sway amplitude of
14.450 over the same runs with a 0.082 standard deviation, and an
integral squared error of 1459.238 degrees squared per second, with a
standard deviation of 8.392 degrees squared per second. For tests at the
same speed with LQR control implemented, we found that over four runs,
the mean max sway amplitude was 4.356 degrees with a standard deviation
of 0. Over the same four runs, the mean 5\% settling time was 2.948
seconds, with a standard deviation of 0.100 seconds. During these runs,
the mean integral squared error was 11.0577 squared degrees per second,
with a standard deviation of 0.007 squared degrees per second.
\begin{table}[t]
\caption{MuJoCo results at 2.96 m/s, four runs per controller, reported
as mean $\pm$ standard deviation. Mean-of-means amplitude was recorded
only for the PD runs and settling time only for the LQR runs, hence the
two empty cells. The LQR reduces peak sway by 70\% and integral squared
error by more than two orders of magnitude.}
\label{tab:mujocoresults}
\centering
\scriptsize
\setlength{\tabcolsep}{4pt}
\begin{tabular}{@{}lcc@{}}
\toprule
Metric & PD wall follower & Six-state LQR \\
\midrule
Mean of means sway amplitude (deg) & $12.505 \pm 0.015$ & N/A \\
Mean max sway amplitude (deg)      & $14.450 \pm 0.082$ & $4.356 \pm 0.000$ \\
Mean 5\% settling time (s)         & N/A                & $2.948 \pm 0.100$ \\
Integral squared error (deg$^2$/s) & $1459.238 \pm 8.392$ & $11.058 \pm 0.007$ \\
\bottomrule
\end{tabular}
\end{table}
\begin{figure}[!htbp]
\centering
\includegraphics[width=\columnwidth]{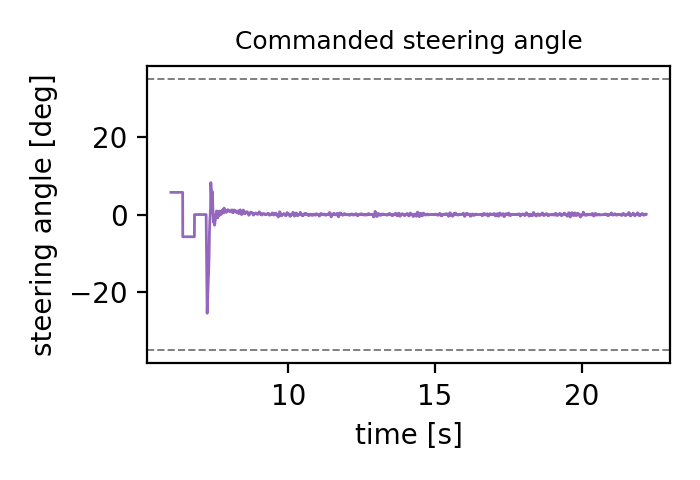}
\caption{Commanded steering angle in the MuJoCo LQR run, with the
$\pm 35$ degree mechanical limit shown dashed. The square steps before
$t = 7$~s are the imposed swerve disturbance; the controller engages
after it, spikes once to about $-26$ degrees, and then holds near zero.}
\label{fig:mujoco_steer}
\end{figure}
The steering control signal graph in Fig.~\ref{fig:mujoco_steer} shows
the controller using a single large correction followed by much smaller
corrections. Peak demand is about $-26$ degrees compared to a $\pm 35$ degree mechanical limit, so the controller never saturates.
\begin{figure}[!htbp]
\centering
\includegraphics[width=\columnwidth]{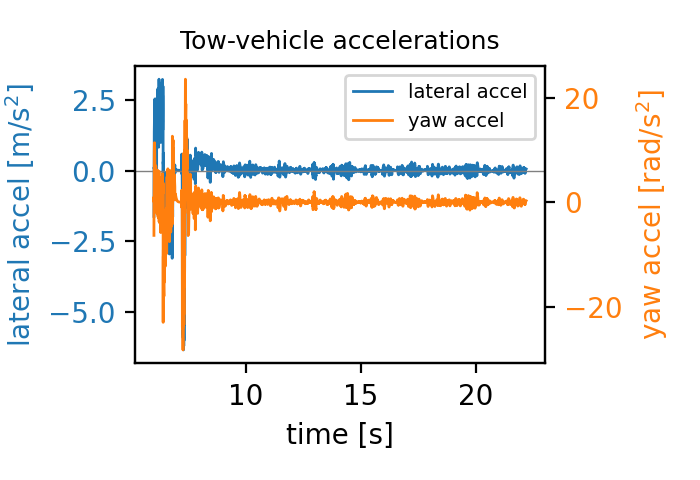}
\caption{Tow-vehicle lateral acceleration (left axis) and yaw
acceleration (right axis) through the MuJoCo LQR run, taken from the car
IMU. Both are large through the disturbance and recovery and fall to a
small residual ripple once the trailer is back in line.}
\label{fig:mujoco_accel}
\end{figure}
Figures~\ref{fig:mujoco_pd} and~\ref{fig:mujoco_lqr} are the same run
under the two controllers, so the panels can be read side by side. Under the PD wall follower the trailer
never settles, and under the six-state LQR it does.
While using the stanard PD wall follower, the  hitch articulation
angle grows from the initial swerve and settles into an oscillation
near $\pm 12$ degrees, the tow vehicle yaws $\pm 35$ degrees relative to the road, and the lane offset reaches $\pm 0.38$~m on every cycle, which
is wider than the 0.60~m lane.
%=====================================================================
\section{Physical Validation}
Next, we tested using the physical Neoracer from Neobotics and our model
of a trailer, which we 3D printed using PLA at 5\% infill. Our data
collection utilized cameras mounted to the car, viewing forwards and
backwards, which captured the front of the car and a view of the trailer.
%=====================================================================
\section{Discussion}
While the MuJoCo simulation and linear model agree on the overall dynamics of the system, several specific numbers were notably different.
The linear  model predicts that rearward cargo placement leads to divergent and then oscillatory behavior, and the MuJoCo does as well: with cargo behind the axle the system sustains a hitch
oscillation under PD wall-following control, visible as a limit cycle in
Fig.~\ref{fig:mujoco_pd}, and the six-state LQR suppresses it within
about 2~s of the disturbance ending, reducing peak amplitude by roughly
70 percent and integral squared error by more than two orders of
magnitude. However, MuJoCo shows sway from 2.68~m/s upward, while
the linear model, run with the cornering stiffnesses currently coded in
the script, does not predict an oscillatory crossing until 5.4~m/s, more than a
factor of three higher.
This is not surprising due to several factors. The linear model  assumes constant normal loads,
whereas the mechanism the abstract identifies is precisely a
\emph{reduction} in rear-axle vertical load; MuJoCo captures that load
transfer, visible in the front-versus-rear axle panel of
Fig.~\ref{fig:mujoco_lqr}, and the linear model cannot. The two models
are also not describing quite the same trailer: the MuJoCo runs place the
cargo 0.040~m behind the axle, while the linear model's ``destabilized''
case has the CG 0.0015~m ahead of it.
The controller result carries over between the two models better than the
stability boundary does, because the same LQR is used.
Despite the 1:12 scale our research focuses on, important core components still apply to a 1:1 scale vehicle. These include the structure of the state-space model, structure of the stability boundaries, and LQR controller architecture.

\subsection{Limitations}
The linear model discards pitch, roll and heave, assumes constant forward
speed, and linearizes the tires, all reasonable near the operating
point and all violated during an actual sway event of any size. The
MuJoCo model relaxes those but introduces its own: rigid tires, a
frictionless hitch, one friction coefficient across all six tires, and
cargo rigidly fixed to the deck. Neither model includes trailer or tow vehicle braking, a deliberate choice, due the the nature of consumer trailers, and the intention of our research.
Additionally, because most of our conclusions and tests were done using a 1:12 scale vehicle (the Neoracer), there are several scale-induced limitations of our findings. Several aspects of the dynamics, such as cornering stiffness, suspension behavior, and aerodynamics don’t scale with other aspects, like mass, inertia, and velocity. Due to this, critical speeds, such as the sway-inducing speed, do not transfer to a 1:1 vehicle. Additionally, state-space model parameters, exact stability boundaries, and controller parameters (including priority weighting and gains) don’t transfer.

\subsection{Future Work}
The immediate next steps are to check the top-of-range gain across a
finer speed sweep than the three points we report, to finish
the physical validation, to identify the cornering stiffnesses so the
analytical and simulated critical speeds can be reconciled, and to log
commanded steering angle and tow-vehicle IMU channels in the MuJoCo CSV
so the remaining figures can be produced. Beyond that, the project document identifies integration
with a SLAM stack as the path to real deployment: the LQR's lane-relative
heading and offset inputs can be replaced with the heading and offset
relative to a planned racing line, which would let the same controller
operate on mapped roads rather than a walled corridor. A comparison
against a model predictive controller, which is what most SLAM stacks
currently use for path following, would establish whether the simplicity
of LQR costs anything material here.
%=====================================================================
\section{Conclusion}
This work developed a linear state-space model of the yaw-plane dynamics
of a light vehicle--trailer combination, extended it with two kinematic
states so that lane keeping and sway damping could be posed as a single
control problem, and synthesized a linear quadratic regulator against it.
Eigenvalue analysis showed that trailer loading does not merely shift the
critical speed but changes which instability mode appears at all:
forward-CG loading fails divergently at 7.5~m/s, in the jackknife mode,
while rearward-CG loading fails oscillatorily at 5.4~m/s, as classical
trailer sway. A controller designed to catch one would not necessarily
catch the other.
The six-state LQR suppresses both without trading sway damping against
lane position, and does so only using front steering, an essential condition since the alternatives options we reviewed
require braking hardware most consumer trailers do not carry. A bounded
Kalman filter estimates trailer states the controller needs from
tractor-only sensing, removing the requirement to instrument the trailer
at all. In MuJoCo, at 2.96~m/s with cargo behind the axle, the controller
reduced peak sway amplitude from 14.450 to 4.356 degrees and integral
squared error from 1459.238 to 11.058 degrees squared per second.
Physical validation on the Neoracer platform remains outstanding.
%=====================================================================
\appendices
\section{Numerical System Matrices}
\label{app:matrices}
All values below are for the destabilized configuration
($a_2 = 0.27866$~m, $I_2 = 0.02107$~kg\,m$^2$) at the design speed
$V = 8.6$~m/s.
\begin{equation}
A_4 =
\begin{bmatrix}
-12.235 & -8.803  &  0.026 & -0.913 \\
 -8.679 & -23.383 & -0.507 & 17.630 \\
-30.148 &  13.828 & -2.035 & 70.703 \\
  0     &   1     & -1     &  0
\end{bmatrix}
\end{equation}
\begin{equation}
B_4 =
\begin{bmatrix} 58.799 \\ 733.002 \\ -293.592 \\ 0 \end{bmatrix}
\qquad
B_6 =
\begin{bmatrix} 58.799 \\ 733.002 \\ -293.592 \\ 0 \\ 0 \\ 0 \end{bmatrix}
\end{equation}
\begin{equation}
M =
\begin{bmatrix}
 9.900 & -0.818 & -1.082 \\
-0.818 &  0.205 &  0.201 \\
-1.082 &  0.201 &  0.286
\end{bmatrix}
\end{equation}
Kalman gain built from the tuned covariances of Table~\ref{tab:cov}
against the same plant:
\begin{equation}
L =
\begin{bmatrix}
    0.171 &  -13.834 &   -0.195 &    0.005 \\
   -1.034 &  143.801 &    0.434 &   -0.103 \\
   -2.097 &  110.290 &   21.973 &   27.112 \\
   -1.821 &  334.735 &    1.068 &    0.853 \\
   -0.029 &    0.863 &    0.414 &    0.531 \\
    0.001 &   -0.280 &    0.727 &    2.535
\end{bmatrix}
\end{equation}
\section{Software}
\label{app:software}
\begin{itemize}
  \item \texttt{swaykalmaanboundedgraph.jl}: Julia. Builds the four-
        and six-state models, sweeps eigenvalues against speed, solves
        the LQR, runs the bounded Nelder--Mead covariance search, and
        generates every plot in the \texttt{plots\_output/} directory.
        Dependencies: \texttt{LinearAlgebra}, \texttt{Roots},
        \texttt{Plots}, \texttt{ControlSystems}, \texttt{Optim}.
  \item \texttt{rc-truck-trailer.xml}: MuJoCo model. CAD-derived
        meshes for the tow vehicle and trailer frame, sprung axles, ball
        hitch with a 45 degree limit, IMUs on both bodies, hitch angle
        sensor, and a 0.60~m lane bounded by walls at $y = \pm 0.95$~m.
  \item \texttt{rc\_sway\_experiment.py}: Python. Regenerates the
        trailer subtree per run for a given cargo offset and mass, runs
        the simulation, applies the controller, simulates the LiDAR,
        logs to CSV, and produces the eight-panel run figures.
  \item \texttt{car\_kalman\_filter.py}: Python. The observer for the
        real vehicle, which unlike the MuJoCo model has no hitch angle
        sensor. Shares its plant with the LQR so the two are always
        designed against the same model.
  \item \texttt{trailer\_model.py}: Python. Holds \texttt{rig\_params},
        \texttt{build\_models} and \texttt{design\_lqr}, and is
        imported by both Python files above so the observer and the
        controller are always designed against the same plant. Include it
        in the submission package.
\end{itemize}
%=====================================================================

\end{document}